\documentclass[sigconf]{acmart}
\AtBeginDocument{%
  \providecommand\BibTeX{{%
    \normalfont B\kern-0.5em{\scshape i\kern-0.25em b}\kern-0.8em\TeX}}}

\copyrightyear{2022}
\acmYear{2022}
\setcopyright{iw3c2w3}
\acmConference[WWW '22] {Proceedings of the ACM Web Conference 2022}{April 25--29, 2022}{Virtual Event, Lyon, France.}
\acmBooktitle{Proceedings of the ACM Web Conference 2022 (WWW '22), April 25--29, 2022, Virtual Event, Lyon, France}
\acmPrice{15.00}
\acmISBN{978-1-4503-9096-5/22/04}
\acmDOI{10.1145/3485447.3512278}

\usepackage{comment}
\usepackage{enumerate}
\usepackage[shortlabels]{enumitem}
\usepackage[singlelinecheck=false,justification=justified]{caption}
\usepackage{algorithm} 
\usepackage{algpseudocode} 
\usepackage{amsmath}
\usepackage{multirow}
\usepackage{float}
\usepackage{lipsum}
\usepackage{mdwlist}
\usepackage{graphics}
\usepackage{tikz,pgfplots}
\usepackage{soul}
\usepackage{capt-of}
\usepackage{caption}
\usepgfplotslibrary{groupplots}
\usepackage{tabularx}

\DeclareMathOperator*{\argmax}{argmax}

\usepackage{kotex}
\usepackage{hyperref}
\hypersetup{ colorlinks=true, linkcolor=blue, filecolor=magenta, urlcolor=cyan, }

\newcommand{\todoblue}[1]{\textcolor{blue}{#1}}

\newcommand{\beq}{\begin{equation}}
	\newcommand{\eeq}{\end{equation}}
\newcommand{\bit}{\begin{itemize*}}
	\newcommand{\eit}{\end{itemize*}}

\definecolor{customblue}{HTML}{20639B}
\definecolor{customred}{HTML}{ED553B}
\definecolor{customyellow}{HTML}{EB9605}
\definecolor{custompurple}{HTML}{9867C5}
\definecolor{customgreen}{HTML}{35B37E}

\begin{document}

% \title[Personalized Draft Recommendation for Victory in MOBA Games]
%       {Personalized Draft Recommendation for Victory 
%       via Hierarchical Player-level and Match-level Representation in MOBA Games}
\title[DraftRec: Personalized Draft Recommendation for Winning in Multi-Player Online Battle Arena Games]{DraftRec: Personalized Draft Recommendation for Winning \\ in Multi-Player Online Battle Arena Games}

\settopmatter{authorsperrow=5}

\author{Hojoon Lee}
\authornote{Equal Contribution. $^\dagger$Work done at KaKaoEnterprise. $^\ddagger$Corresponding author.}
\authornotemark[2]
\email{joonleesky@kaist.ac.kr}
\affiliation{%
  \institution{KAIST}
  %\streetaddress{P.O. Box 1212}
  \country{}
}

\author{Dongyoon Hwang}
\authornotemark[1]
\email{godnpeter@kaist.ac.kr}
\affiliation{
  \institution{KAIST}
  \country{}
}

\author{Hyunseung Kim}
\email{mynsng@kaist.ac.kr}
\affiliation{
  \institution{KAIST}
  \country{}
} 

\author{Byungkun Lee}
\email{byungkun@kaist.ac.kr}
\affiliation{%
  \institution{KAIST}
  \country{}
  %\country{South Korea}
} 

\author{Jaegul Choo}
\authornotemark[3]
\email{jchoo@kaist.ac.kr}
\affiliation{
  \institution{KAIST}
  \country{}
} 

%\author{Hojoon Lee}
%\authornote{Equal Contributions}
%\author{, Dongyoon Hwang*, Hyunseung Kim, Byungkun Lee, and Jaegul Choo}
%\authornotemark[2]
%\email{{joonleesky, godnpeter, mynsng, byungkun.lee, jchoo}@kaist.ac.kr}
%\affiliation{ 
%      \institution{Kim Jaechul Graduate School of AI, KAIST}
%}
%\author{
%\}

\renewcommand{\shortauthors}{Lee et al.}

% \input{tex/00 Abstract}

%%%%%%%%%%%%%%%%%%%%%%%%%
%%%%%% 0. Abstract %%%%%%
%%%%%%%%%%%%%%%%%%%%%%%%%
\begin{abstract}

%  Multiplayer Online Battle Arena (MOBA) games are one of the most popular game genre around the world. When playing MOBA games, players alternatively select a virtual character (a.k.a.champion) to play by not only considering their proficiency, but also the synergy and competence of the team's champion combination. However, current methods for MOBA draft recommendations only considers the winning probability and not the individual characteristics of each players. Without personalization, players would be likely to be less skilled recommended drafts To alleviate this problem, we present DraftRec, a novel sequential recommendation model that recommends champions with a high probability of winning while understanding each player's individual characteristics via player- and \debug{session}-level representations. DraftRec relies on two transformer networks where the former network captures the individual player's preferences and proficiency while the latter network integrates the complex relationship between the players and their respective champion. We train our model with $29,000$ matches of top $0.1\%$ players in \textit{League of Legends} and achieve superior performance compared to various state-of-the-art recommendation methods. 

This paper presents a personalized character recommendation system for Multiplayer Online Battle Arena (MOBA) games which are considered as one of the most popular online video game genres around the world. When playing MOBA games, players go through a \textit{draft} stage, where they alternately select a virtual character to play. When drafting, players select characters by not only considering their character preferences, but also the synergy and competence of their team's character combination. However, the complexity of drafting induces difficulties for beginners to choose the appropriate characters based on the characters of their team while considering their own champion preferences.
To alleviate this problem, we propose DraftRec, a novel hierarchical model which recommends characters by considering each player's champion preferences and the interaction between the players.
DraftRec consists of two networks: the player network and the match network. The player network captures the individual player's champion preference, and the match network integrates the complex relationship between the players and their respective champions. We train and evaluate our model from a manually collected 280,000 matches of \textit{League of Legends} and a publicly available 50,000 matches of \textit{Dota2}. Empirically, our method achieved state-of-the-art performance in character recommendation and match outcome prediction task. Furthermore, a comprehensive user survey confirms that DraftRec provides convincing and satisfying recommendations. Our code and dataset are available at \todoblue{\url{https://github.com/dojeon-ai/DraftRec}}.

%However existing approaches which focus on utilizing search algorithms to recommend the character with the highest winning probability. 

%DraftRec is composed of two networks: (i) policy network, which imitates the expert's drafting procedure, and (ii) value network, which predicts the actual match outcome.
%In deployment, DraftRec recommends the highest valued champion from the value network, whose probability is greater than a certain value of the policy network. This ensures the model to consider both a winning probability and the player's champion proficiency.
%\debug{However, a recommendation that solely relies on the value network can suffer from the distributional shift. Therefore, in deployment, the highest valued champion whose probability is greater than a certain value of the policy network is recommended.}

% DraftRec relies on two transformer networks where the former network captures the individual player's preferences and proficiency while the latter network integrates the complex relationship between the players and their respective champion. 
\end{abstract}



%\ccsdesc[500]{Information systems~Recommender systems}


\maketitle

%%%%%%%%%%%%%%%%%%%%%%%%%%%%%%
%%%%%% 01. Introduction %%%%%%
%%%%%%%%%%%%%%%%%%%%%%%%%%%%%%
\section{Introduction} \label{section:intro}
Multi-player Online Battle Arena (MOBA) games such as \textit{League of Legends} and \textit{Dota2} are one of the most popular online video games. For example, the annual world championship of \textit{League of Legends} was reported to have nearly 50 million viewers in 2020~\cite{LOL2020-onlinearticle}.

%For example, the annual world championship of \textit{League of Legends}' annual World Championship was reported to have nearly 50 million viewers, making it the world's most watched e-sports event in 2020~\cite{LOL2020-onlinearticle}.
When playing MOBA games, multiple players participate in a single session, which we refer to as a \textit{match}. Each \textit{match} is divided into two stages: \textit{draft} stage and \textit{play} stage. In the \textit{draft} stage, players are split into two teams and alternately select a virtual character (i.e., champion). In the following \textit{play} stage, players control their selected champions and the first team that destroys the opponent's main tower wins the game. The \textit{draft} stage is a crucial component in MOBA games since the strategy of the subsequent \textit{play} stage largely depends on the champions selected in the \textit{draft} stage.
%combinations of champions selected in the \textit{draft} stage.

Champions in MOBA games vary in terms of their abilities and required player skills. Thus, it is vital to understand how different champions complement each other (i.e., synergy) and how they counter the abilities of the opponent team's champions (i.e., competence)~\cite{performanceindex}. However, it is challenging to fully understand the synergy and competence since the number of champion combinations are exponential to the total number of champions. For example, in \textit{League of Legends}, there currently exist 156 champions, leading to approximately $4.42 \times 10^{17}$ (i.e., $\binom{156}{5}\times\binom{151}{5}$) possible champion combinations for a single match where there are two teams of five players.
Therefore, players commonly rely on game analytic web services to access information about the relationship between the champions. For instance, the game analytic web service "op.gg"~\cite{LOL2021-OPgg} is accessed by 55 million League of Legends players per month.
%Even} professional MOBA e-sport teams hire coaches to help their team build optimal draft strategies~\cite{LOL2021-coacharticle}.
%In order to resolve these difficulties, players commonly rely on game analytic applications to access information about the relationship between the champions. For example, a widely used \textit{League of Legends}' game analytic web service called "\textit{op.gg}" has more than 45 million monthly visitors worldwide \cite{LOL2021-OPgg}. 
%Due to the complexity of choosing the right champion for a particular situation, players who are unfamiliar with the game are often at a disadvantage even before they start playing the game, which is one of the main factors that makes them lose interest in continuing to play the game.
%Accordingly, the need for a personalized recommender system that relieves the difficulty of selecting champions at the \textit{Draft} stage is increasingly emerging.

To alleviate such difficulties, previous work focused on recommending champions with a high probability of winning by considering the synergy and competence of the champions~\cite{Artofdrafting, ye2020deeprl_moba}. 
However, while matches in MOBA games are  composed of various players with different champion preferences, none of these methods take the player's personal champion preference into consideration.
Therefore, in order to build an effective champion recommendation system, it is essential to consider each player's champion preference and the complex interaction between the players.

To this end, we present DraftRec, a recommender system that suggests champions with a high probability of winning while understanding the champion preference of each player within the match.
To reflect each player's champion preference and then integrate that information, we construct a novel hierarchical architecture composed of two networks: the player network, and the match network. First, the player network captures the individual players' champion preferences. Then, the match network integrates the complex relationship between the players and their selected champions by integrating the outputs from the player network.

The main contributions of this paper can be summarized as follows: (i) we formalize the personalized draft recommendation problem in MOBA games; (ii) we propose DraftRec, a novel hierarchical Transformer-based architecture \cite{vaswani2017transformer} which understands and integrates information about players within a single match; (iii) through comprehensive experiments, DraftRec achieves state-of-the-art performance against personalized recommendation systems in the champion recommendation task and the match outcome prediction task compared to existing MOBA research.
In addition, DraftRec provides effective and satisfying recommendations to the real-world players corroborated by the user study. %conducted on $84$ participants.
%our recommendation strategy of utilizing both the champion prediction and match outcome prediction head enhances user's satisfaction for the draft recommendation.

%%%%%%%%%%%%%%%%%%%%%%%%%%%%%%
%%%%%% 02. Related Work %%%%%%
%%%%%%%%%%%%%%%%%%%%%%%%%%%%%%
\section{Related Work} \label{section:related_work}
\subsection{Recommender Systems}
%Group recommendation section from~\cite{araujo2020itemrec} is worth seeing
% 우리가 결국에 하는 것은 personalized, session-based recommendation이라는 생각이 드넹
% https://lpworld.github.io/files/recsys20.pdf
% 이거 읽기만 해서는 진도 안나간다 우선 복붙 대려!

\noindent
Traditional recommender systems attempt to estimate a user's preferences and recommend items base on them~\cite{adomavicius2005toward}. Such recommender systems are mainly categorized into two groups, content- and collaborative filtering-based recommender systems~\cite{Pazzani2007-contentbased, Koren2011-CF, Sarwar2001-item_CF, Hu2008-Implicit_CF, He2017-ncf, ijcai2017-dmf}. While content-based systems utilize the similarity between items to provide new recommendations, collaborative filtering methods utilize the user's historical feedback to model the degree of matching between users and items. Unfortunately, traditional recommender systems do not consider the temporal nature of human behaviors.

To address these difficulties, deep learning-based sequential recommendation models have been proposed to further exploit the temporal dynamics of user behaviors~\cite{sasrec, bert4rec, GRU4rec, Hidasi2018-RNN_sessionbased,Quadrana2017-HierarchyRNN}. 
%Recently, deep learning-based sequential recommender systems have shown great success
By capturing the complex nonlinear relationship between users and items, recurrent neural networks (RNN)~\cite{Hidasi2018-RNN_sessionbased, Quadrana2017-HierarchyRNN,GRU4rec} were able to explicitly model the sequential nature in user behavior sequences. Due to their great success in natural language processing, deep-learning based recommender systems using attention mechanisms~\cite{sasrec, bert4rec} also have shown promising results in representing sequential data. SASRec~\cite{sasrec} uses a uni-directional self-attention layer from the Transformer's~\cite{vaswani2017transformer} decoder and Bert4Rec~\cite{bert4rec} uses the bi-directional self-attention layer from the Transformer's encoder~\cite{vaswani2017transformer, devlin-etal-2019-bert}.

\subsection{Previous Research in MOBA Games}
\label{subsection:moba_game_research}
Multiplayer Online Battle Arena (MOBA) games are a sub-genre of strategic video games where players form two teams and compete against each other in a virtual battlefield. 
MOBA games have been widely recognized as an ideal test-bed for AI research since the game play dynamics entail complex interactions including cooperation and competition among the players. While MOBA game research has been conducted on a variety of topics such as anomaly detection~\cite{anomalydetection}, player performance evaluation~\cite{performanceindex}, game event prediction~\cite{esportsanalysis, teamfightcamera} and game-play analysis~\cite{Kleinman2020-Gameplay, Cantallops2018-MOBA_review, openai5, Pobiedina2013-Team, ye2020mastering,ye2020deeprl_moba}, our work mainly focus on the followings: (i) devising an accurate match outcome prediction and (ii) a personalized draft recommendation.

Extensive prior research have focused on applying various machine learning methods to properly predict MOBA game match outcomes by utilizing various in-game features~\cite{Artofdrafting, chen2018_synergy, ye2020deeprl_moba, optmatch, glomatch, HOI, livewinprediction}. 
%\debug{While there has been previous work which focuses on producing real-time match outcome predictions utilizing live game information~\cite{livewinprediction}, our work focuses on producing match outcome predictions utilizing the in-game information available at the \textit{draft} stage.} 
HOI~\cite{HOI} predicts match outcomes by computing teammates' pair-wise interactions with a factorized machine based model. OptMatch~\cite{optmatch} and GloMatch~\cite{glomatch} adopts the multi-head self-attention module~\cite{vaswani2017transformer} to predict match outcomes. NeuralAC~\cite{NeuralAC} provides a method which predicts match outcomes by explicitly modeling the synergy and competence between the champions.

%While there has been previous work which focuses on producing real-time match outcome predictions utilizing live game information~\cite{livewinprediction}, our work focuses on producing match outcome predictions utilizing the in-game information available at the \textit{draft} stage such as HOI, OptMatch, GloMatch.

%While there has been previous work which focuses on producing real-time match outcome predictions utilizing live game information~\cite{livewinprediction}, our work focuses on producing match outcome predictions utilizing the in-game information available at the \textit{draft} stage such as HOI, OptMatch, GloMatch.}

Recommendation systems for MOBA games ~\cite{Artofdrafting,lolitemrec} have also been studied, where~\cite{Artofdrafting} recommends champions by leveraging Monte-Carlo Tree Search and~\cite{lolitemrec} recommends \textit{play} stage strategies to increase the probability of victory for the player. However, our work differs with previous work in that they do not take players' personal preferences into account.
%~\cite{lolitemrec} recommends non-personalized \textit{play} stage strategies, while our work provides personalized recommendations for \textit{draft} stage strategies. 

%%%%%%%%%%%%%%%%%%%%%%%%%%%%%%%
%%%%%% 03. Preliminaries %%%%%%
%%%%%%%%%%%%%%%%%%%%%%%%%%%%%%%
\section{Preliminaries} \label{section:preliminaries}
This section provides background information and a formal problem formulation about the drafting process in MOBA games. 

% In Section \ref{subsection:howtodraft}, we explain drafting process in \textit{League of Legends}. In Section \ref{subsection:problem_formulation}, we present the problem formulation of the tasks in two phases: (i) pre-training via supervised learning to learn the player's personal champion selection strategy and (ii) fine-tuning with reinforcement learning to maximize the winning probability.

\subsection{Draft Process in MOBA Games}
\label{subsection:draftprocess}

\begin{figure}[b]
\begin{center}
\includegraphics[width=0.9\linewidth]{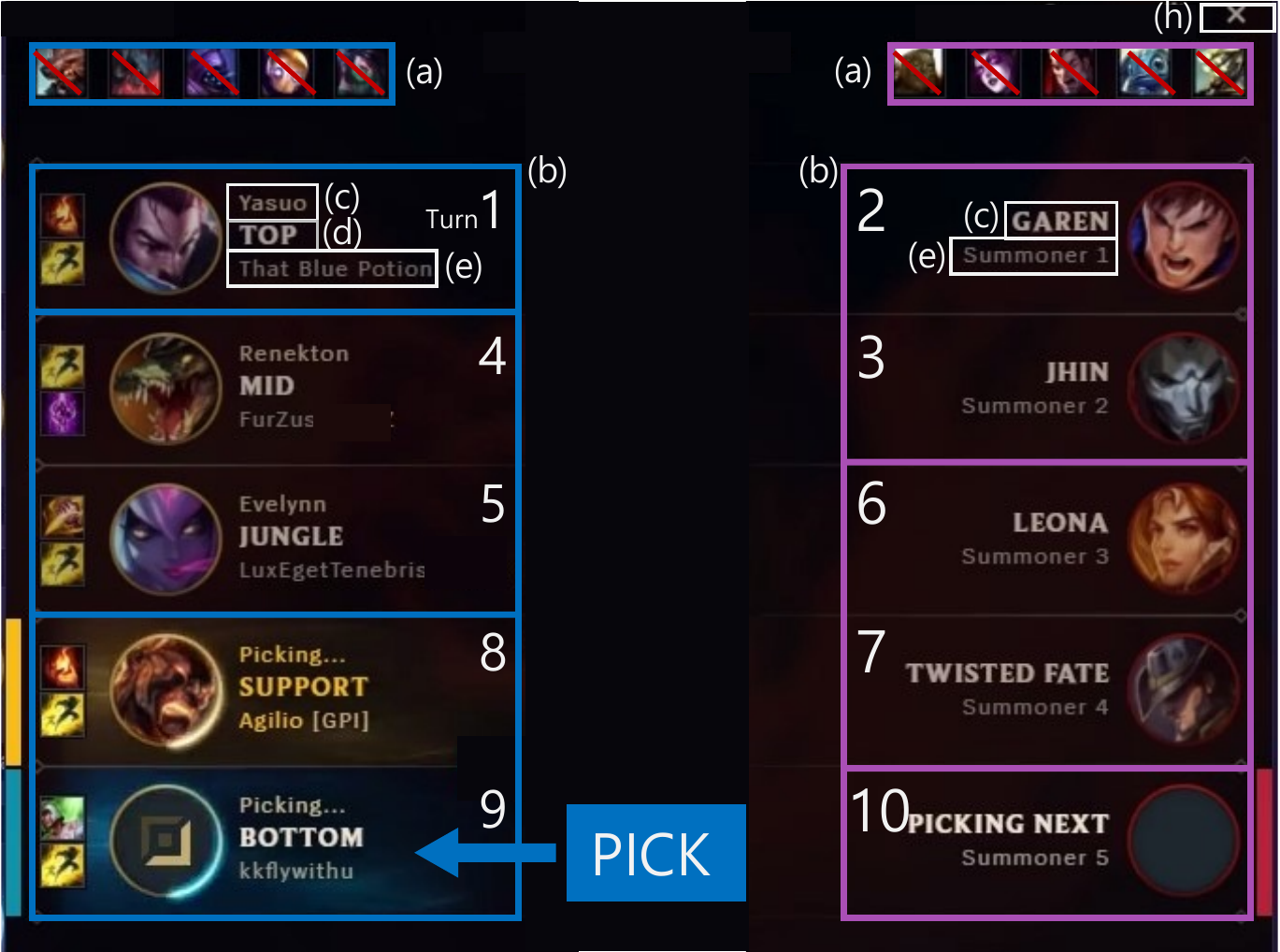}
\end{center}
\captionsetup{justification=centering}
\caption{In-game screenshot of the \textit{draft} stage in \textit{League of Legends}, describing the selection at turn 9. ©Riot Games.}
\label{fig:pick_ban_figure.pdf}
\end{figure}

%Here we provide a detailed explanation of the \textit{draft} stage in MOBA games with an actual example from \textit{League of Legends}. 
Our work focuses on draft processes in competitive modes in MOBA games (e.g., the rank mode for \textit{League of Legends} and the captain's mode for \textit{Dota2}). Since draft processes are similar across different MOBA games, with minor differences, we explain the draft process of MOBA games with an example from \textit{League of Legends}.

As illustrated in Fig.~\ref{fig:pick_ban_figure.pdf}(b), a total of 10 players participate in a single match, where they are divided into two teams: Blue and Purple. Each player is randomly assigned a particular turn (i.e., the order of selecting a champion) and a role (e.g., Top, Jungle, Middle, AD Carry, and Support) as shown in Fig.~\ref{fig:pick_ban_figure.pdf}(b) and (d). After assignment, the ban process follows, where each player chooses champions to be illegal to select for the current match
%, resulting in a total of 10 banned champions
(Fig.~\ref{fig:pick_ban_figure.pdf}(a)). Players then sequentially select champions (Fig.~\ref{fig:pick_ban_figure.pdf}(c)) by considering one's preferences as well as the synergy and competence of their team's champion combination. The champion selection process is always conducted in a team-wise alternating order of "1-2-2-2-2-1" (i.e., 1-(2,3)-(4,5)-(6,7)-(8,9)-10 in Fig.~\ref{fig:pick_ban_figure.pdf}(b)). In selection, previously selected champions (Fig.~\ref{fig:pick_ban_figure.pdf}(c)) are visible to both teams and can no longer be selected by subsequent players within a single match. 

Note that the information available to a particular player in the \textit{draft} stage is partial. That is, as shown in Fig.~\ref{fig:pick_ban_figure.pdf}(d) and (e), the teammates' roles and player IDs are available while the opponents' are not. "Summoner 1" shown in Fig.~\ref{fig:pick_ban_figure.pdf}(e) is an anonymous nickname given to masked player IDs. Using IDs, players commonly search teammates' match histories using game analytic applications to understand teammates' champion preferences.
%, and so does our model, as will be described in Section \ref{subsection:match_network}. %However, if a player judge that their team members and champion compositions have a great disadvantage against the opponent's, they can intentionally abandon (i.e., dodge) the match by closing the current window (Fig.~\ref{fig:pick_ban_figure.pdf}(h)) at the cost of a small loss in ranking compared to the loss in actually losing the game. 

\subsection{Problem Formulation}
\label{subsection:problem_formulation}

The entire drafting process illustrated in Section \ref{subsection:draftprocess} can be formalized as follows. Let us denote a single match as $m^{i}$ (for $1\leq i\leq M$, where $M$ is the total number of matches in our training data) which is composed of 10 champion selections, one per each player. Players $[{p_1^i}, \cdots,{p_{10}^i}]$ (where ${p_t^i}$ represents the player at turn $t$ in match $m^i$) are evenly divided into two teams, $Blue$ $[{p_t^i}]_{t=1,4,5,8,9}$ and $Purple$ $[{p_t^i}]_{t=2,3,6,7,10}$ as shown in Fig.~\ref{fig:pick_ban_figure.pdf}(b). Here, the ground-truth champion, role, match outcome (e.g., win or lose) and team of $p_t^i$ are respectively defined as $c_t^i$, $r_t^i$, $o_t^i$ and $team_t^i$.

% $[c_1^i,..., c_{10}^i], [r_1^i,..., r_{10}^i], [o_1^i,..., o_{10}^i]$, and $[team_1^i,..., team_{10}^i]$. 

% We note that ${p_t}^{i}$ knows the champions selected by turn $t-1$, $[c_k^i]_{k=1:t-1}$, and the roles within the same team, (Fig. \ref{fig:pick_ban_figure.pdf}(d) and (e)).

% along with information about her teammates. 

% We note that ${p_t}^{i}$ has knowledge about the champions selected by turn $t-1$ along with information about her teammates. 
% Given such information, ${p_t}^{i}$ needs to select a champion that she is skilled at and optimizes her team's current draft. 

%%%% history 얘기 %%%%

The \textit{draft} stage of MOBA games can be formulated as a variant of the sequential recommendation problem. The typical sequential recommendation problem aims to predict the player's most preferred champion (i.e., item) based on their champion interaction history~\cite{sasrec, bert4rec}. However, in MOBA games, we have to recommend champions based on not only a single player's champion selection history but also on the teammates' champion selection history. Therefore, we aim to encode each player's champion preference information based on their past champion selection logs as well as their teammates' selection logs.

%While the typical sequential recommendation problem predicts the player's most preferred item based on the user's (i.e., a player)  item (i.e., a champion) interaction history~\cite{sasrec, bert4rec}, in the MOBA game \textit{draft} recommendation setting, we have to provide recommendations based on a single player's champion selection history as well as the history of the teammates'.
%considering the choices and preferences of the other players competing in the same match. 

Let $H_{{p_t^i}, L}$ denote the match history list of player ${p_t^i}$'s most recent $L$ matches before match $m^i$. Then, we denote $c_{{p_t^i}, l},\; r_{{p_t^i}, l}$,  and ${ftr}_{{p_t^i}, l}$ as the champion, role, and list of in-game features (e.g., number of kills, total earned gold, see supplement for feature design) of the $l$-th match in $H_{{p_t^i}, L}$, respectively. Thus, $h_{{p_t^i}, l}$ and $H_{{p_t^i}, L}$ is defined as 
%$c_{{p_t^i}, l},\; r_{{p_t^i}, l}$,  and $o_{{p_t^i}, l}$ be the champion, role, and outcome(i.e., win, lose) of the $l^{th}$ previous match for player ${p_t}^{i}$ at match $m^i$, respectively. Then, $h_{{p_t^i}, l}$ and $H_{{p_t}^{i}, L}$ is defined as 
\begin{equation}
\begin{split}    
    h_{{p_t^i}, l} &= [c_{{p_t^i}, l}, \;r_{{p_t^i}, l}, \;{ftr}_{{p_t^i}, l}] \\
    H_{{p_t^i}, \, L} &= [h_{{p_t^i}, 1}, \;h_{{p_t^i}, 2},\;.\;.\;.\;,\;h_{{p_t^i}, L-1}, \;h_{{p_t^i}, L}]
\end{split}
\end{equation}

\noindent
where $h_{{p_t^i}, L}$ indicates the most recent match history and $h_{{p_t^i}, 1}$ indicates the oldest match history. In addition, we denote the lists $H_{{p_k^i}, \, L}$ and $r_k^i$ of the players in the same team as $H_{team_t^i, \, L}, \; R_{team_t^i}$. 
\begin{equation}
\begin{split}
    H_{team_t^i, \, L} &= \big[ H_{p_k^i, \, L} \; \big| \; team_k^i = team_t^i \, \text{for} \; k=1,...,10 \big] \\
    R_{team_t^i} &= \big[ r_k^i \; \big| \; team_k^i = team_t^i \, \text{for} \; k=1,...,10 \big] 
\end{split}
\end{equation}

\noindent
Then, as described in Section \ref{subsection:draftprocess}, for each champion selection, the players can access currently selected champions, their teammates' match histories, and their respective roles. 
We denote the set of all observable information for the player $p_t^i$ at turn $t$ as state $s_t^i$. 
\begin{equation}
    s_t^i = \{ [c_k^i]_{k=1:t-1}, \;H_{team_t^i, \, L}, \; R_{team_t^i}\}
\end{equation}
%we denote $s_t^i = \{H_{team_t^i, \, L}, \; [c_k^i]_{k=1:t-1}, \; R_{team_t^i}\}$ as the set of all partially observable information for $p_t^i$ at turn $t$. 
Given $s_t^i$, our recommendation model $f_\theta$ outputs two different values, $\hat{p}_t$ and $\hat{v}_t$.
\begin{equation}
    \hat{p}_t^i, \hat{v}_t^i = f_{\theta}(s_t^i)
\end{equation}
For each output $\hat{p}_t$ and $\hat{v}_t$, the goal of the recommendation model is to accurately predict the ground-truth champion $c_t^i$ and the match outcome $o_t^i$ respectively.

%$H_{\mathcal{T}^{(t)}, \,  L}$ be the list of $H_{p_k^i, \, L}$ whose $p_k^i$ is in the same team with $p_t^{i}$

% \text{turn of}p_t^{i}'s\;team\;members} \Big)

%c_0(p_t^{i}) f_\theta \Big( [c_0(p_k^i)]_{k=1:i-1}, \; [S(p_k^i)]_{k=p_t^{i}'s\;team\;members} \Big)
%\textbf{Fine-tuning with  to recommend the champion with the highest chance of winning}
\begin{figure*}[h]
\begin{center}
\includegraphics[width=1.0\linewidth]{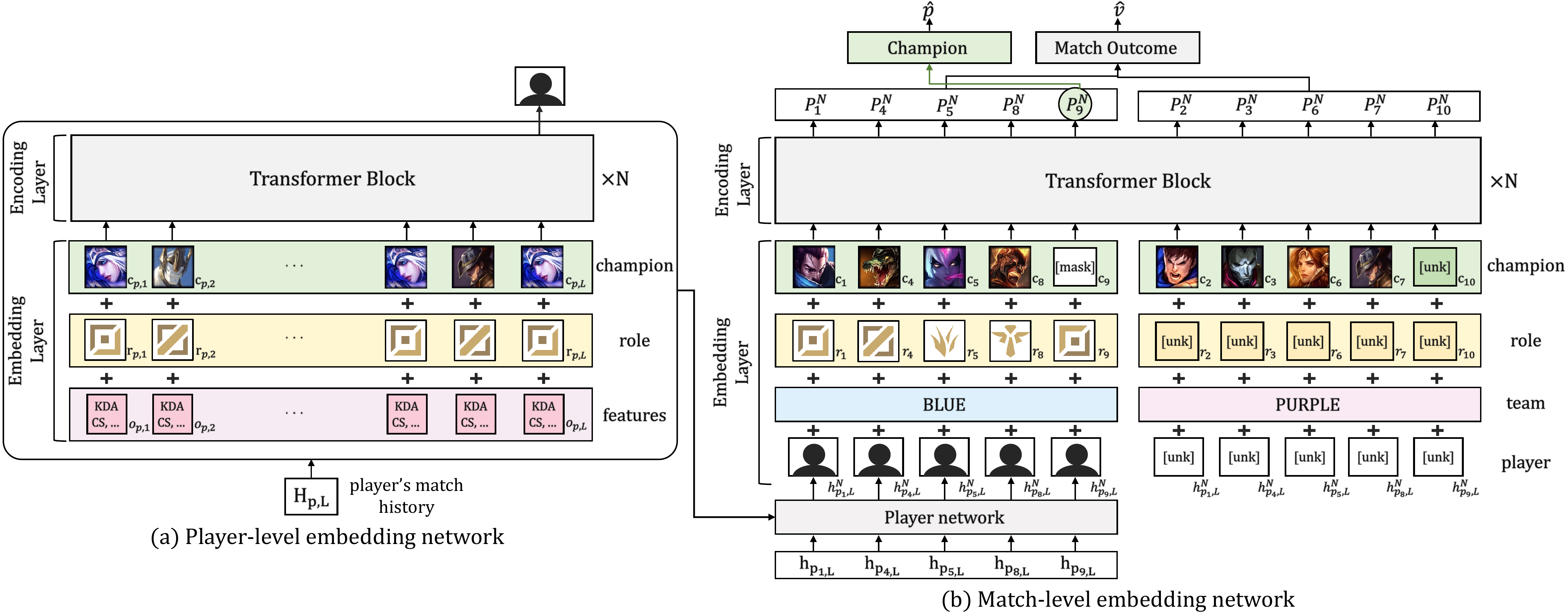}
\end{center}
\captionsetup{justification=centering}
\caption{(a) Player-level embedding network which encodes the player's match history. (b) Match-level embedding network which integrates the player's representation with current match information. For brevity, we omit the positional encoding.}
\label{fig:model}
\end{figure*}

\section{Proposed Method} \label{section:methodology}

This section presents a detailed design of DraftRec. 
DraftRec exploits a hierarchical architecture with the two Transformer-based networks: the player network and the match network. While the player network focuses on capturing the individual players' champion preferences, the followed match network integrates the outputs of the player network which will be explained in Sections \ref{subsection:player_network}-\ref{subsection:match_network}. 
Then, in Sections \ref{subsection:SL_learning_method}-\ref{subsection:recommendation_strategy}, we describe the training procedure and the recommendation strategy of DraftRec.

Throughout this section, we simplify the notation by omitting the index of the match.
To denote the player, champion, role, team, in-game features, match histories, and set of all partially observable information for a player at turn $t$ for match $m^i$, we denote them as $p_t$, $c_t$, $r_t$, $o_t$, $team_t$, $ftr_t$, $H_{p_t, L}$, and $s_t$, respectively. In addition, notations with a superscript will denote a vector embedding instead. 

\subsection{Modeling Player's Dynamic Behavior}
\label{subsection:player_network}

This section presents the details of the player network in Fig.~\ref{fig:model}(a).
%based on their most recent \textit{L} matches.  

\noindent
\\
\textbf{Embedding Layer.}
First, we initialize a learnable embedding for the champion $E_C \in \mathbb{R}^{|C|\; \times \;d}$, role $E_R \in \mathbb{R}^{|R|\; \times \;d}$ and in-game features $E_{FTR} \in \mathbb{R}^{|FTR|\; \times \;d}$, where $d$ indicates the number of hidden units for the embedding. We also initialize a pre-defined sinusoidal positional embedding $E_{pos} \in \mathbb{R}^{|L|\; \times \;d}$ to represent history order.

Given the $l$-th match history $h_{p, l}$ of player $p$, the input embedding $\boldsymbol{h}^0_{p, l}$ is constructed by summing up the corresponding champion, role, in-game feature, and position embedding, i.e., 
%\begin{equation}
%\begin{split}
%    E = E_C^{(j_c)} + E_R^{(j_r)} + E_O^{(j_o)} + E_{pos}^{(n)}\\  
%\end{split}
%\end{equation}
\begin{equation}
\begin{split}
    \boldsymbol{h}^0_{p,l} = \boldsymbol{c}_{p, l}^0 + \boldsymbol{r}_{p, l}^0 + \boldsymbol{ftr}_{p, l}^0 + \boldsymbol{pos}_{l}^0
\end{split}
\end{equation}
where $\boldsymbol{c}_{p, l}^0 \in E_C$, $\boldsymbol{r}_{p, l}^0 \in E_R$, and $\boldsymbol{ftr}_{p, l}^0 \in E_{FTR}$ are $d$-dimensional embeddings for the champion $c_{p, l}$, role $r_{p, l}$, and feature ${ftr}_{p, l}$ while $\boldsymbol{pos}_{l} \in E_{pos}$ represents the positional embedding for position $l$.

Then, the list of $L$ match history $H_{p, L}$ is embedded as
\begin{equation}
\begin{split}
    \boldsymbol{H}^0_{p,L} = [\boldsymbol{h}^0_{p, 1},\; \boldsymbol{h}^0_{p, 2},\;...,\;\boldsymbol{h}^0_{p, 2},\; \boldsymbol{h}^0_{p, L}]
\end{split}
\end{equation}

\noindent
\textbf{Player History Encoder.}
On top of the embedding layer, we place $N$ Transformer blocks $\mathtt{Trm}$, each coupled with a multi-head self-attention network followed by a position-wise feed-forward network~\cite{vaswani2017transformer}. 
To further ease the difficulty of training, we employed a residual connection, layer normalization, and dropout, following \cite{sasrec, bert4rec}. Through passing the Transformer blocks, the list of the embedded match history $\boldsymbol{H}^0_{p,L}$ is transformed as 
\begin{equation}
\begin{split}
   \boldsymbol{H}^n_{p,L} = \mathtt{Trm}(\boldsymbol{H}^{n-1}_{p,L}), \;\; \forall n \in [1, ..., N]
\end{split}
\end{equation}

After $N$ Transformer blocks, the final output ${H}^N_{p,L}$ encodes the information across match histories. Then, the last position of the output $\boldsymbol{h}^N_{p,L}$ is defined as the player-level representation of a player given his/her match history list.
%the embedding of the player-level representation for a given match history list of a player.

\subsection{Modeling Match Information}
\label{subsection:match_network}
This section explains the match network that integrates the outcomes of the player network as illustrated in Fig. \ref{fig:model}(b)

\noindent
\\
\textbf{Embedding Layer.}
Every match $m^i$ is composed of 10 players $[p_1,..., p_{10}]$, their respective roles $[r_1,..., r_{10}]$ and champions $[c_1,...,\\ c_{10}]$, corresponding teams $[team_1,..., team_{10}]$, and the players' past match history lists $[H_{p_1,L},..., H_{p_{10},L}]$. 
At turn $t$, the match information is partially observable for player $p_t$, where only the champions selected before current turn $[c_k]_{k=1:t-1}$, along with the teammates' match histories $H_{team_t, \, L}$ and roles $r_{team_t}$ are known.
Therefore, we replaced the unavailable instances with the "[unk]" token.
Then, we replaced the champion $c_t$ with the "[mask]" token to indicate the query position.
We add an extra learnable embedding $E_{Team} \in \mathbb{R}^{2\; \times \;d}$ to model the players' team information, and initialize the sinusoidal positional embedding  $E_{Turn} \in \mathbb{R}^{|T|\; \times \;d}$ to represent the order of the turn.

Afterward, each player's embedding $\boldsymbol{p}^0_{k}$ at turn $k$ is constructed by summing up the embedding of the corresponding champion, role, team, turn and the output of the player network:
\begin{equation}
\begin{split}
    \boldsymbol{p}^0_{k} = \boldsymbol{c}^0_{k} + \boldsymbol{r}^0_{k} + \boldsymbol{team}^0_{k} + \boldsymbol{turn}^0_{k} + \boldsymbol{h}^N_{p_k,L}
\end{split}
\end{equation}
where $\boldsymbol{c}^0_{k} \in E_C$, $\boldsymbol{r}^0_{k} \in E_R$, $\boldsymbol{team}^0_{k} \in E_{Team}$, and $\boldsymbol{turn}^0_{k} \in E_{Turn}$ are $d$-dimensional embeddings for the champion $c_{k}$, role $r_{k}$, team $team_{k}$, and positional embedding for turn $k$.

Thus, we represent the list of players in match $m^i$ as 
%the following embedding:
\begin{equation}
\begin{split}
    \boldsymbol{P}^0 = [\boldsymbol{p}^0_{1},\;...,\;\boldsymbol{p}^0_{9},\; \boldsymbol{p}^0_{10}]
\end{split}
\end{equation}
%Note that $\boldsymbol{p}^0_{0}$ is the embedding to aggregate the information of the match where $c_0$ is set to a special token "[CLS]" and $\boldsymbol{r}^0_0$, $\boldsymbol{team}^0_0$, $\boldsymbol{turn}_0$, and $\boldsymbol{h}^N_{p_0, L}$ are set to $\boldsymbol{0} \in \mathbb{R}^d$.
\noindent
\textbf{Match Encoder.}
Identical to the player network, we place $N$ Transformer blocks $\mathtt{Trm}$ as encoder of the match network.
For each Transformer block, the list of players in the match is encoded as
\begin{equation}
\begin{split}
   \boldsymbol{P}^n = \mathtt{Trm}(\boldsymbol{P}^{n-1}), \;\; \forall n \in [1, ..., N]
\end{split}
\end{equation}

After $N$ Transformer blocks, the final $d$-dimensional output ${P}^N$ serves as the match representation where the personal histories of the players and available information for each turn of the match are aggregated. 

\noindent
\\
\textbf{Champion Prediction Head.}
We now predict the ground-truth champion $c_t$ based on the hierarchically integrated player representation $p^N_t$ at turn $t$. 
In particular, two fully-connected layers with a GELU nonlinear activation is utilized where a softmax layer is followed to produce an output probability distribution over target champions as $\hat{p}_t \in \mathbb{R}^{|C|}$:
\begin{equation}
    \hat{p}_t = \text{softmax}(GELU(\boldsymbol{p}^N_t W^P + b^P)E_C^T + b^C)
\end{equation}
where $W^P \in \mathbb{R}^{d \times d}$ is the learnable projection matrix and $b^P$, $b^C$ are bias terms. 
To prevent the banned champions from being recommended, predicted scores of banned champions are masked out.

\noindent
\\
\textbf{Match-Outcome Prediction Head.}
We jointly perform the match outcome prediction by comparing the representations of the two teams. Similar to \cite{optmatch, glomatch, NeuralAC}, we obtain a $d$-dimensional representation vector of each team $ {T_1},  {T_2}$ by applying the average pooling operation for the player representations within each team. 
\begin{equation}
     {T}_1 = \text{pool}(\{\boldsymbol{p}_1^N,..., \boldsymbol{p}_9^N\}),  \;\;
     {T}_2 = \text{pool}(\{\boldsymbol{p}_2^N,..., \boldsymbol{p}_{10}^N\})
\end{equation}

Then, we subtract the two team representations and apply two fully-connected layers followed by a sigmoid function to obtain the match outcome prediction value $\hat{v}_t$ such as
\begin{equation}
    \hat{v}_t = \text{sigmoid}((({T}_1-{T}_2) W^{O} + b^{O})W^{V} + b^{V})
\end{equation}
where $W^O \in \mathbb{R}^{d \times d}$ and $W^V \in \mathbb{R}^{d \times 1}$ are a learnable projection matrix and $b^O$ and $b^V$ are bias terms. Note that since the match outcome prediction value $\hat{v}_t$ is squashed to [0,1], it indicates the predicted probability of winning. 

\subsection{Training DraftRec}
\label{subsection:SL_learning_method}

This section explains the supervised training procedure of DraftRec. 
For each training match data $m$ at turn $t$, we first get state $s_t$, which represents all partially observable information in respect to player $p_t$. 
Then, as illustrated in Section \ref{subsection:match_network}, we obtain the predicted champion selection probability and predicted match outcome by forwarding the state $s_t$ into the DraftRec model $f_{\theta}$.
\begin{equation}
    \hat{p}_t, \hat{v}_t = f_{\theta}(s_t)
\end{equation}
Then, the network parameters $\theta$ are trained to maximize the predicted champion selection probability $\hat{p}_t$ of the ground-truth champion $c_t$ while minimizing the error between the predicted match outcome $\hat{v}_t$ and the ground-truth match outcome $o_t$.

Here, we denote $\mathcal{L}_p$ as the champion prediction loss and $\mathcal{L}_v$ as the match outcome prediction loss. We use binary-cross entropy for both $\mathcal{L}_p$ and $\mathcal{L}_v$ where L2 weight regularisation is utilized to avoid overfitting. In summary, loss $\mathcal{L}$ for each match is written as
\begin{equation}
    \mathcal{L} = \frac{1}{T}\sum_{t=1}^{T} \; \lambda \mathcal{L}_p(\hat{p}_t^{i} , \;\hat{c}_t^{i}) + 
    (1 - \lambda) \mathcal{L}_v(\hat{v}_t^{i} , \;\hat{o}_t^{i}) + c||\theta||^2
\end{equation}
where $\lambda$ controls the strength between the champion prediction loss and match outcome prediction loss, and $c$ controls the level of the L2 weight regularisation.

\subsection{Recommendation Strategy}

As described in Section \ref{subsection:SL_learning_method}, our model can predict the champion selection probability  $\hat{p}_t$ and match outcome $\hat{v}_t$ for a given state $s_t$. Here, we denote the champion recommended by our model as $\hat{c}$ and the probability of selecting an arbitrary champion $c$ as $\hat{p}_{t,c}$.
To estimate the match outcome of playing the champion $c$ at a given state $s_t$, we first modify the state $s_t$ and fill the "[mask]" token in position $c_t$ with champion $c$. Then, we replace the champion $c_{t+1}$ with a "[mask]" token. Next, we pass the modified state to the model and obtain the predicted match outcome which we denote as $\hat{v}_{t,c}$.

Since we can predict the expected winning probability $\hat{v}_{t,c}$ for all champions $c$, the straightforward recommendation strategy is to recommend the champion with the highest winning probability which already takes the player's champion preferences into account.
\begin{equation}
    \hat{c} = \argmax_c{\hat{v}_{t,c}}
\end{equation}

\label{subsection:recommendation_strategy}
\begin{figure}[b]
\begin{center}
\includegraphics[width=0.97\linewidth]{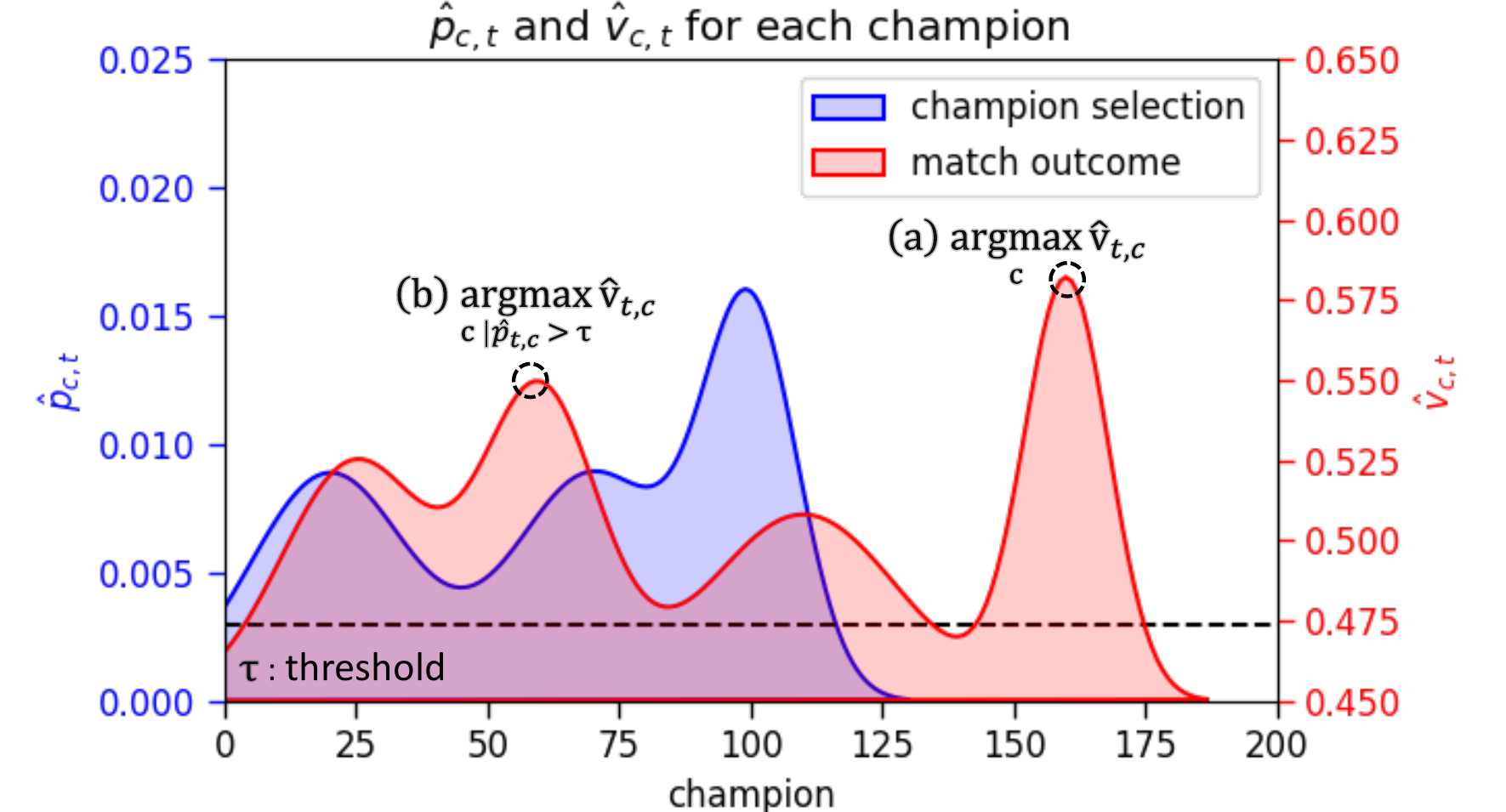}
\end{center}
\captionsetup{justification=centering}
\caption{Illustration of different strategies. (a) indicates the champion with the highest win rate. (b) indicates the highest win rate champion among the preferred ones.} 
\label{fig:recommender_distribution}
\end{figure}

However, a recommendation system which solely rely on the match outcome prediction can exhibit unreliable behaviors. 
Fig. \ref{fig:recommender_distribution} demonstrates an illustrative example where the blue and red curve represents the distribution of $\hat{p}_{t,c}$ and $\hat{v}_{t,c}$ respectively. 
Since the states that are unseen within the training data might have arbitrarily inaccurate prediction values \cite{hendrycks2016baseline, liang2018enhancing}, the highest match outcome value can be inappropriately assigned to champions which players do not prefer, such as positions indicated at Fig.~\ref{fig:recommender_distribution}(a).

Therefore, to properly integrate the player's champion preferences and expected winning probability, our model recommends the champion with the highest match outcome where its selection probability exceeds a threshold value $\tau$ as in Fig \ref{fig:recommender_distribution}.(b).
\begin{equation}
    \hat{c} = \argmax_{c|\;\hat{p}_{t,c} > \tau}{\hat{v}_{t,c}}
\label{equation:draftrec}
\end{equation}

Recently, this idea of restricting the decision space within the training distribution has been actively studied to deploy safe decision-making systems \cite{fujimoto2019off, kumar2019stabilizing, wu2019behavior}. 
In various real-world applications (e.g., recommender systems and robotics), these restrictions have shown reliable and robust results which are also corroborated by our experimental findings in Section \ref{rec_strategy_eval}.

%%%%%%%%%%%%%%%%%%%%%%%%%%%%%%
%%%%%% 05. Experiments %%%%%%%
%%%%%%%%%%%%%%%%%%%%%%%%%%%%%%

\section{Experiments} \label{section:experiments}
This section presents the experimental setup, experimental results, and a detailed description of our user study.

\subsection{Dataset}
\label{subsection:dataset}

To verify the performance of DraftRec, we conduct experiments on two MOBA game datasets: \textit{League of Legends} (LOL) and \textit{Dota2}. For each benchmark dataset, we sort the matches by time-stamps and take the first $85\%$ matches as training set, next $5\%$ matches as validation set, and last $10\%$ matches as test set. The statistics for each dataset are given in Table \ref{table:dataset}.
\vspace{-0.15cm}
\begin{table}[h]
\caption{Statistics of the benchmark datasets.}
\begin{tabular}{l c c c c c}
\toprule
Dataset & Matches & Champions & Players & Avg. Match History \\ \hline
LOL     & 279,893  & 156    & 62,466   & 66.38               \\ 
%TODO: Unkown user 처리하기
Dota2   & 50,000   & 111    & 140,931  & 2.12                \\
\bottomrule
\end{tabular}
\label{table:dataset}
\end{table}

\vspace{-0.15cm}
\noindent\textbf{League of Legends} We manually collected match data for \textit{League of Legends} utilizing the publicly accessible API endpoint provided by Riot Games~\cite{riot_api_link} and constructed a MOBA game match dataset with rich individual player history. To ensure the quality of each match outcome, matches of the top $0.1\%$ ranked players from June 1, 2021, to September 9, 2021 were collected. Since the purpose of building a \textit{draft} recommender system is to provide strategically advantageous suggestions, it is natural to train the model with matches from top rank players since they better understand the characteristics of champions compared to low rank players.

\noindent
\\
\textbf{Dota2} Since previous work on champion recommendation for MOBA games~\cite{Artofdrafting} utilized a Dota2 dataset to validate the performance of their recommendation system, we also used a publicly available Dota2 dataset from Kaggle~\cite{kaggle_dota},  which was collected from November 5, 2015 to November 18, 2015. The dataset contains matches from various ranks and the average match history length of each player is short, as shown in Table \ref{table:dataset}.

%%%%%%%%%%%%%%%%%%%%%%%%%%%%%%%%%%%%%%%%%%
%%%%%% 5.1 Supervised Pre-training %%%%%%%
%%%%%%%%%%%%%%%%%%%%%%%%%%%%%%%%%%%%%%%%%%

\begin{table*}[h]
\begin{center}
%{\small
\caption{Performance comparison of DraftRec and baselines on champion recommendation. Bold scores indicate the best model and underlined scores indicate the second best. The results are averaged over 10 random seeds.}
\begin{tabular}{l c c c c c c c c c c c c c}
\toprule
&\\[-3ex]
\multirow{2}{*}{Models} 
& \multicolumn{5}{c}{LOL} & \multicolumn{5}{c}{Dota2}  \\[-0.5ex]
\cmidrule(lr){2-6} 
\cmidrule(lr){7-11} 
                           & HR@1    & HR@5   & NG@5   &HR@10   & NG@10  
                           & HR@1    & HR@5   & NG@5   &HR@10   & NG@10  \\[0.2ex]
\hline \\[-2.4ex]
POP                        & 0.3212  & 0.5646 & 0.4497 & 0.6553  & 0.4792 
                           & 0.0508  & 0.1131 & 0.0824 & 0.1647  & 0.0989 \\ 
NCF~\cite{He2017-ncf}      & 0.1376  & 0.4219 & 0.2823 & 0.5900  & 0.3367 
                           & 0.0403  & 0.1384 & 0.0899 & 0.2407  & 0.1226 \\ 
DMF~\cite{ijcai2017-dmf}   & 0.3243  & 0.5758 & 0.4567 & 0.6742  & 0.4887 
                           & \textbf{0.0562}  & \underline{0.1488} & \textbf{0.1030} & 0.2302  & 0.1290 \\ 
S-POP~\cite{GRU4rec}       & 0.4135  & 0.7404 & 0.5865 & 0.8469  & 0.6213 
                           & \underline{0.0554}  
                           & 0.1253 & 0.0901 
                           & 0.1773 & 0.1004  \\ 
SASRec~\cite{sasrec}       & \underline{0.4497} & \underline{0.7430} & \underline{0.6071} & \underline{0.8547} & \underline{0.6368} 
                           & 0.0449  
                           & 0.1477 & 0.0951       
                           & \underline{0.2542} & \underline{0.1296} \\ [0.2ex]
\hline \\[-2.4ex]
DraftRec-no-history        & 0.1456  & 0.3950 & 0.2711 & 0.5675 & 0.3268   
                           & 0.0403  & 0.1390 & 0.0905 & 0.2438 & 0.1239 \\ 
DraftRec                   & \textbf{0.5042*} 
                           & \textbf{0.8025*} & \textbf{0.6646*} 
                           & \textbf{0.8836*} & \textbf{0.6618*} 
                           & 0.0434  
                           & \textbf{0.1492} & \underline{0.0962} 
                           & \textbf{0.2547} & \textbf{0.1496*} \\ 
\bottomrule \\[-2.0ex]
\multicolumn{5}{l}{Notes: "\textasteriskcentered" indicates the statistical significance (i.e., $p < 0.01$).}
\end{tabular}
\label{table:recommendation}
\end{center}
\end{table*}

\subsection{Experimental Setup} \label{subsection:experimental_setup}
\noindent
\textbf{Baselines.}
To verify the recommendation performance, we compare our model with the personalized recommendation baselines:
\vspace{-0.05cm}
\renewcommand\labelitemi{\tiny$\bullet$}
\begin{itemize}[leftmargin=*]
\item POP : It is the simplest baseline which ranks item based on the frequency within the given player history. 
\item NCF~\cite{He2017-ncf} : It captures the nonlinear interactions between players and items through a MLP with implicit feedback. 
\item DMF~\cite{ijcai2017-dmf} : It optimizes the Latent Factor Model based on the explicit item selection ratio of each user.
%feedback using cosine similarity between the low dimensional vectors of players and items.
\item S-POP~\cite{GRU4rec} : It is a variant of POP that ranks items based on the player's most recent $n$ history.
\item SASRec~\cite{sasrec} : It utilizes the uni-directional Transformer structure for modeling the player's preference over time.
\end{itemize}

\noindent
For the match outcome prediction, we compare our model with:
\vspace{-0.05cm}
\renewcommand\labelitemi{\tiny$\bullet$}
\begin{itemize}[leftmargin=*]
\item Majority Class (MC)~\cite{Artofdrafting} : It is the simplest baseline which outputs the majority class (i.e., Blue for \textit{LOL} and Radiant for \textit{Dota2}). 
\item Logistic Regression (LR)~\cite{NIPS2001_LR} : It is a linear classifier with L2 regularization. We use the identical input format as \cite{semenov2016performance}.
\item Neural Network (NN)~\cite{Artofdrafting} : It is the neural network model which follows the architecture and input format as \cite{Artofdrafting}.
\item HOI~\cite{HOI} : It is based on factorization machines~\cite{Rendle2010_FM} and considers pair-wise interactions between players.
\item OptMatch~\cite{optmatch} : It exploits graph neural networks to obtain hero embeddings which are used to model players' champion preferences and proficiency. It further utilizes multi-head self attention to obtain team representations from the player representations.
\item NeuralAC~\cite{NeuralAC} : It is a self attentive method which explicitly models the synergy and competence of different champions. However, it doesn't utilize any in-game features or players' match histories.
\end{itemize}

\noindent
\\
\textbf{Training Details.}
\label{subsubsection:training_details}
We re-implement the aforementioned baselines by strictly following the details of each paper. For a fair comparison, we set the number of layers $N=2$ and used the Adam optimizer with $\beta_1=0.9$, $\beta_2=0.999$ with gradient clipping when its $l_2$ norm exceeds $5$. We fix the dimension of each head as $64$ if the models utilized the self-attention mechanism. 

For all models, we tune the common hyper-parameters where learning rate $lr$ ranges from $\{0.01, 0.001, 0.0001\}$ and anneal with a cosine scheduler. We vary the $l_2$ regularization from $\{0.01, 0.001, 0.0001, \\ 0.00001\}$, dropout rate $p$ from $\{0, 0.1, 0.2\}$, hidden dimension size $d$ from  $\{64, 128, 256\}$, and batch size from $\{256, 512\}$. For the models which use match histories (i.e., S-POP, SASRec, NeuralAC, and DraftRec), we vary the match history length $L$ from $\{0, 1, 10, 20, 50, 100\}$. 

%\debug{Since Dota2 does not have pre-defined roles, we exclude such information for our Dota2 dataset experiments.}

Since Dota2 does not have pre-defined roles, we do not utilize the role information when experimenting on the Dota2 dataset.

\noindent
\textbf{Evaluation.}
To evaluate the performance of the champion recommendation and match outcome prediction tasks, we measure the metric of each state $s_t$ at turn $t$ for all matches in the test dataset. Then, we report the average value of each metric.

%we adopted the next item recommendation task, commonly used in sequential recommendation literature~\cite{sasrec, bert4rec}. 
%For each , we utilized previous interaction histories as input to 
%verify whether the model recommends the ground-truth champion.
We employ standard recommendation metrics, \textit{Hit Ratio} (HR), and \textit{Normalized Cumulative Gain} (NG), to evaluate the quality of champion recommendations. We report HR and NG with varying the rank $k$ from $\{1, 5, 10\}$. Since NG@1 is identical to HR@1, we omit NG@1. For match outcome prediction task, we consider Accuracy (ACC) and Mean Absolute Error (MAE) as our metrics. For all metrics except MAE, higher value indicates better performance.

\vspace{-0.15cm}
\subsection{Experimental Results} \label{subsection:experiment_results}
In this section, we study the performance of all methods on the champion recommendation and match outcome prediction task. 

\vspace{-0.15cm}
\noindent
\\
\textbf{Champion Recommendation.} Table \ref{table:recommendation} summarizes the performance of all models on the champion recommendation task. From our experimental results, we can observe the followings:
All sequential methods (i.e., S-POP, SASRec, and DraftRec) outperform non-sequential methods (POP, NCF, and DMF) on all metrics except for HR@1 and NG@5 for the \textit{Dota2} dataset. 
This result shows that dynamically modelling players' preferences improves champion recommendation performance. 

Among all models, DraftRec achieved the best recommendation performance for all metrics and datasets except for HR@1 and NG@5 in Dota2. We speculate that this is due to the sparse player history record in the Dota2 dataset, making it difficult to learn the champion preference of each individual player. 
%For \textit{League of Legends}, DraftRec shows an improvement of $12.1\%$ in HR@1, $8\%$ in HR@5, $9.5\%$ in NG@5 compared to SASRec, and for \textit{Dota2}, DraftRec shows an improvement of $15.4\%$ in NG@10 compared to SASRec.
Compared to SASRec, DraftRec shows an improvement of $12.1\%$ in HR@1, $8\%$ in HR@5, $9.5\%$ in NG@5, $3.4\%$ in HR@10 and $3.9\%$ in NG@10 
for \textit{League of Legends}. For \textit{Dota2}, DraftRec shows an improvement of $15.4\%$ in NG@10 against SASRec. 
In addition, when we trained DraftRec without any player match history information, the recommendation performance degrades by a significant margin. This confirms that integrating both player- and match-level representations is crucial in providing personalized champion recommendations.
%the current state-of-the-art method. By comparing DraftRec with SASRec, we show that integrating the learned player-level representations to the match-level representation is essential in modeling the drafting procedure. 
%On the other hand, without any player match history, we find that DraftRec's performance degrades by a significant margin. Therefore, we find that modeling both player and match-level representations is crucial in providing personalized draft recommendations.
%Comparing DraftRec with DraftRec-no-history, we find that utilizing the game-specific information (e.g., role, team) during the drafting process is beneficial in predicting the next champion. 
%Moreover, personalized modeling is also an important factor since BERT4Rec and DMF are superior to S-POP and POP respectively. 

\begin{table}[h]
\begin{center}
\caption{Performance comparison of DraftRec and baselines on the match outcome prediction task.}
\begin{tabular}{l c c c c c}
\toprule
&\\[-3ex]
\multirow{2}{*}{Models} 
& \multicolumn{2}{c}{LOL} & \multicolumn{2}{c}{Dota2}  \\[-0.5ex]
\cmidrule(lr){2-3} 
\cmidrule(lr){4-5} 
                              & ACC    & MAE    & ACC  & MAE  \\[0.2ex]
\hline \\[-2.0ex]
MC~\cite{Artofdrafting}       & 0.5040 & 0.4960 & 0.5180 & 0.4820 \\ 
LR~\cite{NIPS2001_LR}         & 0.5255 & 0.4973 & 0.5750 & 0.4819 \\ 
NN~\cite{Artofdrafting}       & 0.5263 & 0.4975 & 0.5748 & 0.4822 \\
HOI~\cite{HOI}                & 0.5264 & 0.4987 & 0.5716 & 0.4901 \\ 
OptMatch~\cite{optmatch}      & \underline{0.5411} & 0.4944 & \underline{0.5751} & 0.4842 \\
NeuralAC~\cite{NeuralAC}      & 0.5266 & 0.4977 & 0.5739 & 0.4841 \\ 
 [0.2ex]
\hline \\[-2.0ex]
DraftRec-no-history           & 0.5284 &\underline{0.4942} & 0.5745 & \textbf{0.4757*}  \\ 
DraftRec                      & \textbf{0.5535*} & \textbf{0.4842*} 
                              & \textbf{0.5755} & \underline{0.4782} \\ 
\bottomrule \\[-2.0ex]
\multicolumn{5}{l}{Notes: "\textasteriskcentered" indicates the statistical significance (i.e., $p < 0.01$).}
\end{tabular}
\label{table:match_outcome}
\end{center}
\end{table}

\noindent
\\
\textbf{Match Outcome Prediction.} Table \ref{table:match_outcome} summarizes the performance of all match outcome prediction baselines. We find that for all datasets, match outcome prediction methods which utilize player match history information (i.e., OptMatch, DraftRec) show superior performance compared to methods which do not (i.e., LR, NN, HOI, NeuralAC, DraftRec-no-history). This demonstrates the importance of integrating the player's match histories in order to understand the dynamics behind the match outcome between the players. 

In addition, our proposed DraftRec outperforms all compared baselines for all metrics and datasets. For \textit{League of Legends}, DraftRec shows an improvement of $2.3\%$ in ACC and $2.1\%$ in MAE. For \textit{Dota2}, DraftRec shows an improvement of $1.3\%$ in MAE. Identical to the findings from the champion recommendation task, we observe that utilizing both player-level and match-level representations is beneficial for the match outcome prediction task. For further experimental results of predicting the match outcome only at the post-draft stage (i.e., after the draft is completed), see Appendix~\ref{subsection:appendix_match_outcome_prediction}.

\subsection{Interpreting the Attention Weights} \label{subsection:attention_weight_analysis}
\label{attention_weights_analysis}
After training DraftRec, we analyzed the model's attention weights on \textit{League of Legends} dataset to understand how the model learned the synergy and competence between the different roles of players.

\begin{figure}[b]
\begin{center}
\includegraphics[width=0.8\linewidth]{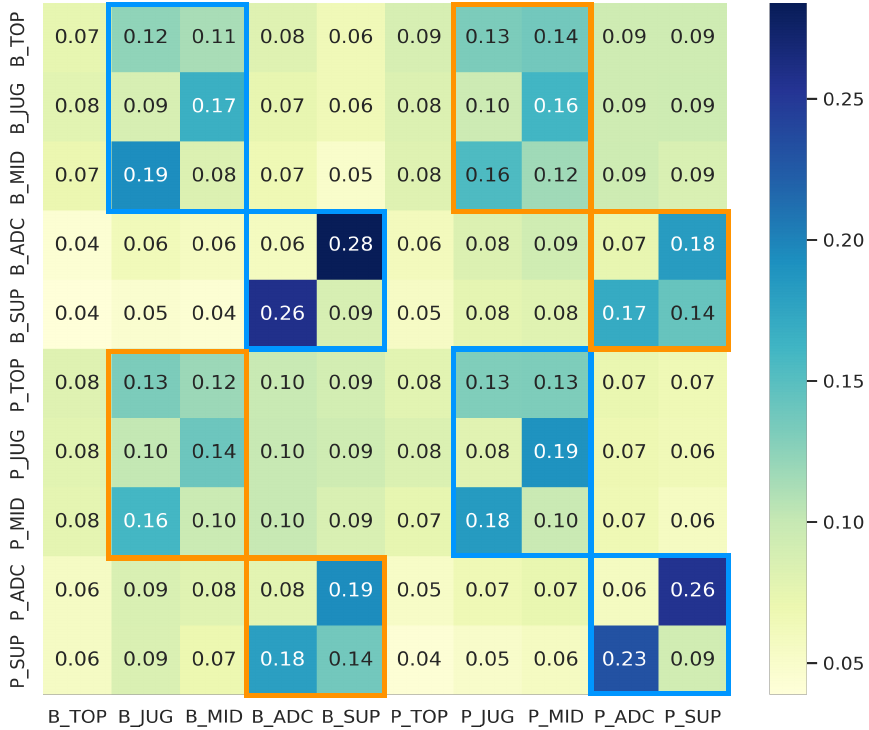}
\end{center}
%\captionsetup{justification=centering}
\caption{Visualisation of the attention weights on \textit{League of Legends}. The weights are ordered by the teams and roles.}
\label{fig:attention_weights}
\end{figure}

Fig.~\ref{fig:attention_weights} displays the average attention weights for the final self-attention layer of the match network over the test dataset. 
Blue (Orange) boxes show high attention weights within the same (different) team.
%Blue boxes show high attention weights within the same team. Orange boxes indicate high attention weights between different teams. 
From Fig.~\ref{fig:attention_weights}, we observe that champions with the Top and Middle roles have high attention scores between champions with the Jungle role. In addition, champions with the AD Carry and Support roles show high attention scores with each other. 
Interestingly, this interaction reflects the actual role interaction of 
%how players with different roles interact with each other when actually playing \textit{League of Legends}. 
\textit{League of Legends}, where (Top, Jungle), (Middle, Jungle), and (AD Carry, Support) mainly interact with each other in a match. We conclude from the above analysis that DraftRec learns meaningful relationships between the players according to their role. 

%To evaluate whether the attention distribution of DraftRec is reasonable, 
%We visualize the average  attention weight of the test set for \textit{League of Legends} matches at Figure \ref{fig:attention_weights}. 
%Typically, in \textit{League of Legends}, champions with Top, Jungle, and Mid roles are important at the beginning of the game while champions with AD Carry, and Support role are more important at the end of the game. Thus, when playing \textit{League of Legends}, the following role pairs have a high correlation: (Top, Jungle, Mid) and (AD Carry, Support).  In Figure \ref{fig:attention_weights}, we are able to observe that (Top, Jungle, Mid) pair has high attention scores with each other and that (AD Carry, Support) pair has a high attention score with each other. Surprisingly, they do not only have high attention scores within the same teams, but also have high attention scores for the opponent team with the same roles. We believe that the model was able to learn the most likable role for each champion.

%In Figure \ref{fig:attention_map},

\subsection{Evaluation for Recommendation Strategy}
\label{subsection:recommendation_strategy_evaluation}
\label{rec_strategy_eval}

%In Section \ref{rl_result}, we give a detailed explanation of the champion recommendation strategy DraftRec utilizes to output champion recommendations the player prefers to play and has a high probability of winning.
%For Section \ref{player_study}, we explain the procedure and present the results of the user study conducted to evaluate the quality of our method's recommendations 

Here, we study the effectiveness of our recommendation strategy which recommends the champion with the highest winning probability among the champions preferred by the player. 
For comparison, we consider three different recommendation strategies.

\noindent
\renewcommand\labelitemi{\tiny$\bullet$}
\begin{itemize}[leftmargin=*]
\item DraftRec$_p$: Recommendation strategy which ranks the champions based on the champion selection probability $\hat{p}_{t,c}$. 
\item DraftRec$_v$: Recommendation strategy which ranks the champions based on the match outcome prediction $\hat{v}_{t,c}$.
\item DraftRec$_{p+v}$: Our proposed recommendation strategy described in Equation \ref{equation:draftrec}. We set the selection probability threshold $\tau$ as $0.02$ (i.e., $1\over50$) to allow the model to recommend champions that were selected within the most recent $50$ previous matches.
\end{itemize}
% 위에서 언급한것 처럼 우리의 모델은 predicted probability of champion p와 expectecd match outcome v_{t,c}를 얻을 수 있다.
% To compare different recommendation strategy,  

%While a straightforward evaluation method to compare the different strategies is online A/B testing, it would be prohibitively expensive and may
%hurt players' in-game experiences. 
To compare the different strategies, a straightforward evaluation method is online A/B testing. However, this could be prohibitively expensive and may hurt players' in-game experiences. 
Therefore, we utilize an offline evaluation method which directly estimates the results with a separate evaluation model \cite{gulcehre2020rl, voloshin2019empirical, zou2020pseudo, xiao2021general}.

For each turn in all test match data, DraftRec recommends a champion according to its corresponding recommendation strategy and a separate match outcome evaluation model predicts the match outcome of the draft assuming that the recommended champion is selected. 
If we use an identical model for both of our recommendation and match outcome evaluation, the correlation of inaccurate predictions can be problematic \cite{wang2019mbcal}. Therefore, we use OptMatch \cite{optmatch} as our separate match outcome evaluation model.
We report HR@10, NG@10, and the average predicted win rate (Win) with varying rank $k$ from $\{3, 10\}$ to evaluate each strategy.

\begin{table}[h]
\caption{Performance comparison of different recommendation strategies evaluated on the \textit{League of Legends} dataset.}
\begin{tabular}{l c c c c}
\toprule
Model                            &HR@10             & NG@10            & Win@3              & Win@10        \\
\hline
DraftRec$_{p}$                   & \textbf{0.8836}  & \textbf{0.6618}  & \underline{52.72}  & \underline{52.46} \\ 
DraftRec$_{v}$                   & 0.2051           & 0.0938           & 51.34              & 51.55              \\
DraftRec$_{p+v}$                 & \underline{0.8495} & \underline{0.5657} & \textbf{54.12} & \textbf{52.59}    \\
\bottomrule
\end{tabular}
\label{table:rec_strategy}
\end{table}

\begin{table*}[h]
\caption{Results of the user survey (N = 84). Bold indicates the best rank for each question and underlined indicates the second.}
%\resizebox{\textwidth}{!}
%{
\begin{tabular}{l c c c c c} %{20mm} {20mm} {20mm} {20mm} {20mm}}
\toprule
\textbf{Rank of the Players}   
&\textbf{All} &\textbf{Diamond$\uparrow$} &\textbf{Platinum}&\textbf{Gold}&\textbf{Silver$\downarrow$} \\

%\textbf{Percentile of each Rank}   
                                                                &$0$-$100\%$    &$0$-$1\%$         &$1$-$10\%$         &$10$-$40\%$           &$40$-$100\%$\\ 
\textbf{Question}                                               &(N=84)         &(N=14)            &(N=23)             &(N=28)                &(N=19)\\
\hline
Q1. Are you familiar with the model's recommended champions?    &86.90\%        &\textbf{100\%}    & \underline{91.3\%}          &85.71\%   &73.68\%\\
Q2. Do you think recommended champions will lead to victory?    &79.76\%        &\textbf{92.86\%}         &\underline{86.96\%}            & 71.43\%  & 73.68\% \\
Q3. Are the explanations of the recommendations reasonable?     &80.95\%        &\textbf{85.71\%}          &73.91\%  & \underline{85.71\%}            & 78.95\% \\
Q4. Do you wish to use this recommender system afterward?       &86.90\%        &\textbf{92.86\%}          &86.96\%  &  \underline{89.29\%}           &78.95\% \\
\bottomrule
\end{tabular}
\label{table:survey_result}
\end{table*}

\vspace{-0.2cm}

Table \ref{table:rec_strategy} summarizes the performance of different recommendation strategies on the \textit{League of Legends} dataset. Our proposed recommendation strategy, DraftRec$_{p+v}$ has achieved the highest predicted match outcome value with a win rate of $54.12\%$ when one of the top-3 recommended champions from the model is selected. DraftRec$_{v}$ shows the worst performance in all metrics where HR@10 and NG@10 drop significantly compared to DraftRec$_{p}$. 
We suspect that the poor performance of DraftRec$_{v}$ is incurred from assigning inaccurate values to the champions outside of the training data distribution as illustrated in Section \ref{subsection:recommendation_strategy}. 

%Since the predictions that are outside of the training data distribution might have arbitrarily inaccurate values \cite{hendrycks2016baseline, liang2018enhancing}, the highest match outcome value can be inappropriately assigned to champions which players do not prefer as Fig \ref{fig:recommender_distribution}.(a).

%From our experimental results, we can observe the followings.
%While DraftRec$_{RL}$ with $\lambda=1.0$ shows the best REW@1 and REW@3 results, it also shows the worst HR@1, HR@3 and NDCG@3 result among the different $\lambda$ values. However, these low HR@1, HR@3, NDCG@3 metric values are not problematic but rather intended. During the fine-tuning phase, DraftRec$_{RL}$'s objective is to learn to predict the champion that gives the highest winning probability, not the champion that the player actually played, which naturally degrades the performance of HR@1, HR@3, and NDCG@3.

%Since our objective is to provide personalized, but yet winnable champion recommendation, 
%we've select DraftRec$_{RL}$ with $\lambda=0.9$ as our final model for the user study which shows reasonable personalization performances (e.g. HR@1, HR@3, NDCG@3) while having a small reward (e.g. REW@1, REW@3) difference to the DraftRec$_{RL}$ with $\lambda=1.0$. 

\vspace{-0.15cm}
\subsection{User Study} \label{subsection:user_study}
To further evaluate our model's recommendations, we conducted a user study with a visualization of DraftRec's recommendations.

\begin{figure}[h]
\includegraphics[width=0.95\linewidth]{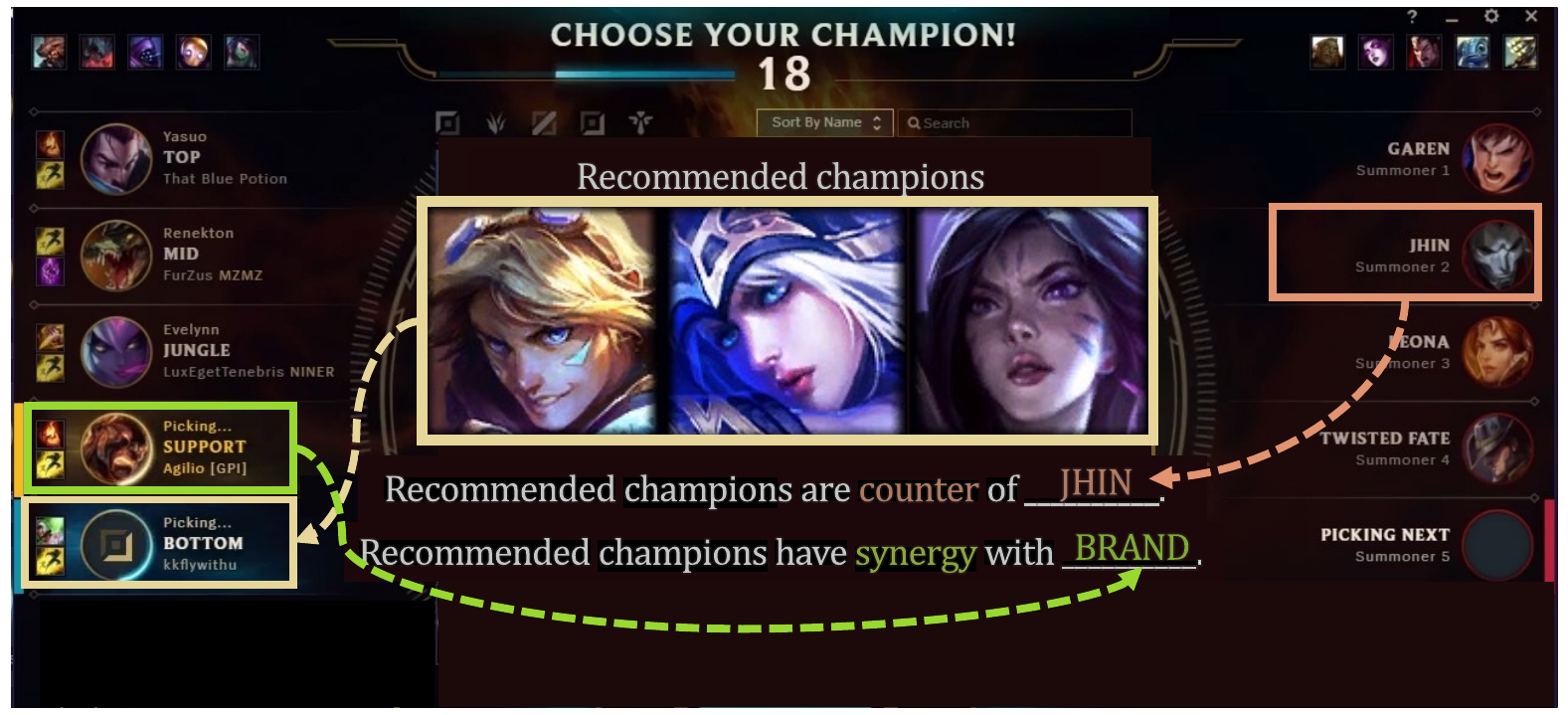}
\captionsetup{justification=centering}
\caption{A user study screen which recommends top 3 champions to the 9-th user. ©Riot Games.}
\label{fig:survey_screen}
\end{figure}

%\textbf{Interpreting the attention mechanism} 
%\label{subsection:attention}
%As illustrated in Fig.~\ref{fig:attention_figure.pdf}. (b), we visualize the attention weights of DraftRec$_{RL}$ in order to interpret the recommendation strategy. It consists of: (i) the selected champions till the current turn $t$, on the left and bottom, and (ii) a heat map in the center. The row where the champion is masked on the left represents the current champion selection turn's attention weights. For example, in Fig.~\ref{fig:attention_figure.pdf}. (b), the current turn is $t=9$. The heat map displays the importance of each champion upon each turn using various gray-scale color intensities. Darker color indicates higher importance. When the attention weight is high for an opponent turn, we interpret our recommendation to be the \textit{counter} champion against that turn's champion and when the attention weight is high for a turn on the same team, interpret that turn's champion as a \textit{synergy} champion. For example in Fig.~\ref{fig:attention_figure.pdf}. (b), the model is currently paying attention to turn $t=3$ and $t=8$ and since our current turn $t=9$, the opponent $t=3$'s champion is the counter champion and the same team $t=8$'s champion is the synergy champion.

\vspace{-0.15cm}
\noindent
\textbf{The user study procedure.}
A total of $84$ user study participants were gathered from various popular online \textit{League of Legends} communities~\cite{invenglobal, talkopgg}. When  gathering  players  to  participate  in  our user study, we explained that we are currently developing a recommendation system trained on top $0.1\%$ ranked user data and that we aim to recommend personalized champions with a high winning probability. To conduct a personalized survey, we first received consent and then collected the account IDs of the participants. Then we gathered the 50 most recent match data of each player and reproduced the actual draft situation they experienced. For each match, the participants were given a display of 3 champion recommendations and a description made based on the attention weights, as shown in Fig.~\ref{fig:survey_screen}. The \textit{synergy} (\textit{counter}) champion indicates the champion with the highest attention value within the player's (opponent's) team.
%and the \textit{counter} champion indicates the champion with the highest attention value within the opponent's team. 
User study participants were given personalized surveys made based on their match history data. 

\begin{comment}
Participants answered the 4 questions listed in Table \ref{table:survey_result}.

We designed the questions as such to answer the following research questions:
\begin{itemize}[leftmargin=*]
\item \textbf{Q1}: Does DraftRec produce personalized recommendations?

\item \textbf{Q2}: Does DraftRec take into account the victory of the game?

\item \textbf{Q3}: Are the learned attention weights of DraftRec reasonable?

\item \textbf{Q4}: Does DraftRec meet actual players' needs?
\end{itemize}
\end{comment}

\vspace{-0.2cm}
\noindent
\\
\textbf{User study results.} 
%$84$ people participated in our user study. We divided the participants into groups based on their \textit{League of Legends} ranks. $14$ out of $84$ participants were in the top $1\%$ rank (Diamond and higher), $23$ were in the top $1\%$-$10\%$ rank (Platinum), $28$ were in the top $10\%$-$40\%$ rank (Gold), and $19$ were in top $40\%$-$100\%$ rank (Silver and lower). 
The survey took an average of 30 minutes for each player. The survey results are presented in Table \ref{table:survey_result}. In terms of \textbf{Q1}, we observe that participants are familiar with the recommended champions ($All=86.9\%$). With respect to \textbf{Q2}, we observe most participants think that our recommendation will lead to victory ($All=79.76\%$), but this perception is more positive to high ranked users ($Diamond=92.86\%$). With regard to \textbf{Q3}, we can see that participants found the explanations for the recommendations reasonable ($All=80.95\%$). Finally, in \textbf{Q4}, we notice that a great majority of the players wish to use our recommender system in the future ($All=86.9\%$). Interestingly, we were able to see a tendency that our method was more positive to high ranked players compared to low ranked players. We speculate that this is due to the difference in the understanding of MOBA games players have according to their rank. 
Another possible explanation is that high ranked players related more with the recommendations since our model was trained with the top $0.1\%$ rank match data.
%Additionally, high ranked players were more satisfied with our model since it was trained on top $0.1\%$ rank match data.
%since the model is trained with the top $0.1\%$ ranked players match data, 
%high ranked players were more satisfied with our model than low ranked players.

%\textbf{Reinforcement Signal ($\lambda$)} In Table \ref{table:lambda_comparison}, we analyze our model's behavior as the strength of the reinforcement signal vary.  

\vspace{-0.1cm}
\section{Ethical Consideration}
\label{section:ethical_consideration}
%In order to ensure that there are no privacy issues within our datasets, all datasets used in this paper went through a preprocessing step for the purpose of anonymizing or removing all personal information. 
We took careful steps to preserve the ethics of research.
%and to ensure that there are no privacy issues.
%All datasets utilized in this paper were thoroughly checked to ensure that there are no privacy or ethical issues.
Specifically, the collected \textit{League of Legends} dataset utilized the publicly accessible API endpoint provided by Riot Games~\cite{riot_api_link} which includes an automatic account ID encryption stage, preventing any abuse of personal information. For the \textit{Dota2} dataset, we utilized a public dataset where all personal information were removed. 
In addition, the user study in Section~\ref{subsection:user_study} was approved by the Research Ethical Committee of KAIST.

%%%%%%%%%%%%%%%%%%%%%%%%%%%%%%
%%%%%% 06. Conclusion %%%%%%%
%%%%%%%%%%%%%%%%%%%%%%%%%%%%%%
\vspace{-0.1cm}
\section{Conclusion and future work} \label{section:conclusion}
In this paper, we present DraftRec, a novel recommendation system which understands each player's champion preference and the complex interaction between different players within a match.
DraftRec utilizes a distinctive strategy to recommend the champions with a high win rate among the preferred ones of each player.
Through extensive experiments on two MOBA-game datasets, we empirically demonstrate the superiority of DraftRec over various baselines and through a comprehensive user study, find that DraftRec provides satisfactory recommendations to real-world players.

A limitation of our work is that while in-game features are dependent on the performance of other players in a match, we only consider them from a single player's perspective. Expanding our method to further integrate other players' performance when constructing the players' match history is left for future work. We hope that our research will contribute to enhanced user experience for various MOBA-game-related web services and applications.

%\debug{One limitation of this work is that we only consider a match record from a single player's perspective when constructing the match histories while in-game features are also dependent on the other players in the same match. Another limitation of this work is that it is currently geared towards high-ranked players. Expanding our method to further integrate other players' characteristics when constructing the players' match history and to account for players in various ranks will be an exciting new direction to explore. We will make our code and manually collected dataset public, and hope that our research will contribute to enhanced user experience for various MOBA game related web services and applications.}

%Our algorithm can be further extended by integrating the other players’ characteristics in constructing the players' match history. Also, our current work is limited in that it is geared towards high-ranked players. Expanding this to players in various ranks will be an exciting direction to explore.} 

%for both the personalized champion recommendation and match outcome prediction task. To further verify the effectiveness of our model, we conducted a comprehensive user study with $84$ participants gathered from various MOBA game communities and achieved high satisfactory results. We will make our code and manually collected dataset public, for the reproducibility of our work. We hope this work will encourage future MOBA game research.
\vspace{-0.1cm}
\section*{Acknowledgments}
This work was supported by Institute of Information \& communications Technology Planning $\&$ Evaluation (IITP) grant funded by the Korea government(MSIT) (No. 2019-0-00075, Artificial Intelligence Graduate School Program(KAIST), and No. 2020-0-00368, A NeuralSymbolic Model for Knowledge Acquisition and
Inference Techniques).

\clearpage

\bibliographystyle{ACM-Reference-Format}
\bibliography{reference}
\clearpage

\appendix

\section{Implementation Details}

Here, we describe the implementation details of DraftRec.

\vspace{-0.1cm}
\subsection{Feature Selection}
\label{subsection:appendix_feature_selection}
Table~\ref{table:selected_features_lol} and Table~\ref{table:selected_features_dota2} lists the selected features for the \textit{League of Legends} dataset and \textit{Dota2} dataset respectively.
We selected features that express the characteristic and champion preference of each player through a careful discussion with 5 top 1\% ranked players of each game. 

\vspace{-0.1cm}
\begin{table}[h]
\caption{Selected features for \textit{League of Legends}.}
 \begin{tabular}{p{4cm}| p{4cm}}
 \toprule
 Feature & Description \\
 \hline
 KDA  & (\#kills + \#assist) / \#death \\ %(\#kills + \#assist) / \#death) 
 Largest\_killing\_spree & Largest killing spree  \\ 
 Largest\_multi\_kill  & Larget number of multi kills \\
 Killing\_sprees & Number of consecutive kills \\
 Longest\_time\_spent\_living & Longest time spent alive \\ 
 Double\_kills & Number of double kills \\ 
 Triple\_kills & Number of triple kills \\
 Quadra\_kills & Number of quadra kills \\
 Penta\_kills & Number of penta kills \\
 Unreal\_kills & Number of unreal kills \\
 Total\_damage\_dealt & Total damage dealt \\
 Magic\_damage\_dealt & Magic damage dealt \\
 Physical\_damage\_dealt & Physical damage dealt \\
 True\_damage\_dealt & True damage dealt \\
 Largest\_critical\_strike & Largest critical strike damage \\
 Total\_damage\_to\_champions & Total damage to champions \\
 Magic\_damage\_to\_champions & Magic damage to champions \\ 
 Physical\_damage\_to\_champions & Physical damage to champions \\ 
 True\_damage\_to\_champions & True damage to champions \\
 Total\_heal & Total heal amount\\
 Total\_units\_healed & Total number of units healds \\ 
 Damage\_self\_mitigated & Damage self mitigated \\ 
 Damage\_dealt\_to\_objectives & Damage dealt to objectives \\
 Damage\_dealt\_to\_turrets & Damage dealt to turrets \\
 Vision\_score & Vision score \\
 Time\_ccing\_others & Time crowd controlling others \\
 Total\_damage\_taken & Total damage taken \\
 Magical\_damage\_taken & Magical damage taken \\
 Physical\_damage\_taken & Physical damage taken \\
 True\_damage\_taken & True damage taken \\
 Gold\_earned & Total gold earned \\
 Gold\_spent & Total gold spent \\
 Turret\_kills & Number of turret kills \\ 
 Inhibitor\_kills& Number of inhibitor kills \\ 
 Total\_minion\_killed& Total number of minions killed \\
 Minions\_killed& Minions killed by player \\
 Minions\_killed\_team\_jungle & Team jungles killed by player \\ 
 Minions\_killed\_enemy\_jungle & Enemy jungles killed by player \\ 
 Total\_time\_crowd\_control    & Time being crowd controlled \\
 Vision\_wards\_bought & Number of vision wards bought \\
 Sight\_wards\_bought  & Number of sight wards bought \\
 Wards\_placed & Number of wards placed \\
 Wards\_killed & Number of wards killed \\
 \bottomrule
 \end{tabular}
\captionsetup{justification=centering}
\label{table:selected_features_lol}
\end{table}

\begin{table}[h]
\caption{Selected features for \textit{Dota2}.}
 \begin{tabular}{p{4cm}| p{4cm}}
 \toprule
 Feature & Description \\
 \hline
 Kills & Number of kills\\ 
 Deaths & Number of deaths \\ 
 Assists & Number of assists \\
 Denies & Number of denies \\
 Xp\_hero & Experience obtained by heroes \\ 
 Xp\_creep & Experience obtained by creeps\\ 
 Xp\_roshan & Experience obtained by roshan\\
 Xp\_per\_min & Experience obtained per minute\\
 Xp\_other & Experience obtained by other sources\\
 Level & The hero's level\\
 Hero\_damage & Damage done by hero \\
 Hero\_healing & Heal amount done by hero \\
 Tower\_damage & Total tower damage\\
 Last\_hits & Number of last hits\\
 Stuns & Number of stuns \\
 Gold & Total amount of gold earned\\
 Gold\_spent & Total amount of gold spent\\ 
 Gold\_per\_min & Gold earned per minute\\ 
 Gold\_other &  Gold earned by other sources\\
 Gold\_death & Gold lost by dying \\
 Gold\_buyback & Gold spent on buybacks\\ 
 Gold\_abandon & Gold earned by abandonment \\ 
 Gold\_sell & Gold earned by selling items\\
 Gold\_destroying\_structure & Gold earned by structures\\
 Gold\_killing\_heroes & Gold earned by killing heroes\\
 Gold\_killing\_creeps & Gold earned by creeps\\
 Gold\_killing\_roshan & Gold earned by roshan\\
 \bottomrule
 \end{tabular}
\captionsetup{justification=centering}
\label{table:selected_features_dota2}
\end{table}

\subsection{Hyperparameters}
\label{subsection:appendix_hyperparameters}

Table~\ref{table:hyperparameters} describes the optimal hyper parameter configurations of DraftRec.
All DraftRec models are trained using A100 GPU with 64 cores. For \textit{League of Legends}, training takes $5$ hours to complete and for \textit{Dota2}, training takes $1$ hour to complete.

\begin{table}[h]
\caption{Hyperparameters of DraftRec on \textit{League of Legends} and \textit{Dota2} dataset.}
 \begin{tabular}{p{3.5cm}|p{2cm}|p{2cm}}
 \toprule
 Hyperparameter             & LOL           & Dota2 \\
 \hline
 Epoch                      & 10            & 20 \\ 
 Optimizer                  & Adam          & Adam \\ 
 Adam $\epsilon$            & 1e-8          & 1e-8 \\
 Adam ($\beta_1, \beta_2$)  & (0.9, 0.999)  & (0.9, 0.999) \\
 LR Scheduler               & Cosine        & Cosine \\
 Initial LR                 & 1e-3          & 1e-3 \\
 Final LR                   & 0             & 0 \\
 Weight decay               & 1e-5          & 1e-4 \\ 
 Batch size                 & 512           & 512 \\
 Hidden size                & 128           & 64 \\
 Clip gradients             & 5.0           & 5.0 \\
 Hidden dropout             & 0.1           & 0.2 \\
 Attention dropout          & 0.1           & 0.2 \\
 Num heads                  & 2             & 1 \\
 Num blocks $N$             & 2             & 1 \\
 Max sequence length $L$    & 50            & 20 \\
 Loss control $\lambda$     & 0.1           & 0.1 \\
 Selection threshold $\tau$ & 0.02          & - \\
 \bottomrule
 \end{tabular}
\captionsetup{justification=centering}
\label{table:hyperparameters}
\end{table}

\section{Further Experiments}

\subsection{Match Outcome Prediction}
\label{subsection:appendix_match_outcome_prediction}

%The conventional setting for training a match outcome prediction model in MOBA games is to train the model to predict the match outcome after the drafts are completed (i.e., every player has selected a champion). 
In our draft recommendation setting, the model needs to predict the match outcome at every turn for each match. Therefore, we trained and evaluated the match outcome prediction models with the partially observable information set at each turn. However, previous match outcome prediction research \cite{Artofdrafting, NIPS2001_LR, HOI, optmatch, NeuralAC} focused on predicting the match outcome when a draft was finished (i.e., prediction was made after every player has selected a champion). Therefore, in Table \ref{table:fin_match_outcome}, we report the performance of all models when they are trained and evaluated by following the experimental protocols of previous work.

\begin{table}[h]
\begin{center}
\caption{Performance comparison of DraftRec and baselines on the match outcome prediction task when predictions are made after every player completed selecting a champion.}
\begin{tabular}{l c c c c c}
\toprule
&\\[-3ex]
\multirow{2}{*}{Models} 
& \multicolumn{2}{c}{LOL} & \multicolumn{2}{c}{Dota2}  \\
\cmidrule(lr){2-3} 
\cmidrule(lr){4-5} 
                             & ACC    & MAE    & ACC    & MAE  \\[0.2ex]
\hline \\[-2.0ex]
MC~\cite{Artofdrafting}      & 0.5040 & 0.4960 & 0.5180 & 0.4820 \\ 
LR~\cite{NIPS2001_LR}        & 0.5323 & 0.4975 & \textbf{0.6126} & 0.4682 \\ 
NN~\cite{Artofdrafting}      & 0.5335 & 0.4958 & 0.6108 & 0.4692\\
HOI~\cite{HOI}               & 0.5339 & 0.4978 & 0.6076 & 0.4819 \\ 
OptMatch~\cite{optmatch}     & \underline{0.5449} & 0.4928 & 0.6109 & 0.4724 \\
NeuralAC~\cite{NeuralAC}     & 0.5347 & 0.4962 & 0.6108 & 0.4712 \\ 
 [0.2ex]
\hline \\[-2.0ex]
DraftRec-no-history          & 0.5432 & \underline{0.4893} & 0.6085 & \textbf{0.4592*}  \\ 
DraftRec                     & \textbf{0.5618*} & \textbf{0.4826*} & \underline{0.6110} & \underline{0.4642} \\ 
\bottomrule \\[-2.0ex]
\multicolumn{5}{l}{Notes: "\textasteriskcentered" indicates the statistical significance (i.e., $p < 0.01$).}
\end{tabular}
\label{table:fin_match_outcome}
\end{center}
\end{table}

Similar to the findings of our match outcome prediction experiments in Section \ref{subsection:experiment_results}, DraftRec performs the best among all methods for the \textit{League of Legends} dataset. Compared to OptMatch~\cite{optmatch}, which is a state-of-the-art match outcome prediction model, DraftRec shows an improvement of $3.1\%$ in ACC and $2.1\%$ in MAE. 
However, for the \textit{Dota2} dataset, we observed that a simple logistic regression model performed the best. 
% Dota2에서는 simple한 logistic regression model이 성능이 제일 좋았는데 this implies that 
% 이는 player의 정보가 충분히 고려되지 않을 경우 

\subsection{Ablation on Maximum Sequence Length} \label{subsection:appendix_ablation_study}

In order to further analyze the importance of utilizing personal match histories, we conduct an ablation studies referring to the length of the player match history the models utilizes. Since \textit{Dota2} dataset contains a scarce amount of players' match histories, we only consider \textit{League of Legends} dataset throughout this experiment. 
For all the experiments, we vary the length of player match histories $L$ from $\{1,2,5,10,20,50,100\}$ and fix the remaining hyper-parameters to each model's optimal configurations.

\hspace{-0.55cm}
\pgfplotsset{compat=1.16}
\begin{tikzpicture}
\begin{groupplot}[
      group style={group size=2 by 2, horizontal sep=1.2cm, vertical sep=1.2cm},
      width=4.7cm, height=4.7cm]
    %------------------------------------------------------------%
    %\nextgroupplot[],
    \nextgroupplot[
        every x tick label/.append style={font=\scriptsize\color{gray!80!black}},
        xmin=0.8, xmax=130,
        xtick={1,2,5,10,20,50,100},
        xticklabels={1,2,5,10,20,50,100},
        xlabel style={yshift=1mm},
        xmode=log,
        %log ticks with fixed point,
        xlabel={\footnotesize$L$},   
        every y tick label/.append style={font=\scriptsize\color{gray!80!black}},
        ymin=0.38, ymax=0.92,
        ylabel={\footnotesize HR@10},
        ylabel style={yshift=-1mm},
        grid=both,
        tick align=outside,
        tick pos=left,
        legend style={at={($(0,0)+(1cm,1cm)$)},
        legend columns=4,fill=none,draw=black,anchor=center,align=center, font=\small},
        legend to name = models]
    \addplot[color=customblue,
             mark=*,
             mark size=1] 
    coordinates{(1, 0.775) (2, 0.815) (5, 0.867) (10, 0.872) (20, 0.878) (50, 0.883) (100, 0.887)};
    \addplot[color=customyellow,
             mark=*,
             mark size=1.1] 
    coordinates{(1, 0.405) (2, 0.506) (5, 0.643) (10, 0.730) (20, 0.802) (50, 0.841) (100, 0.847)};
    \addplot[color=customred,
             mark=*,
             mark size=1.1] 
    coordinates{(1, 0.505) (2, 0.565) (5, 0.675) (10, 0.754) (20, 0.825) (50, 0.846) (100, 0.854)};
    \addlegendentry{DraftRec};    
    \addlegendentry{S-POP};
    \addlegendentry{SASRec};
    \coordinate (c1) at (rel axis cs:1,1);
    \nextgroupplot[
        every x tick label/.append style={font=\scriptsize\color{gray!80!black}},
        xmin=0.8, xmax=130,
        xtick={1,2,5,10,20,50,100},
        xticklabels={1,2,5,10,20,50,100},
        xlabel style={yshift=1mm},
        xmode=log,
        %log ticks with fixed point,
        xlabel={\footnotesize$L$},   
        every y tick label/.append style={font=\scriptsize\color{gray!80!black}},
        ymin=0.35, ymax=0.7,
        ylabel={\footnotesize NG@10},
        ylabel style={yshift=-1mm},
        grid=both,
        tick align=outside,
        tick pos=left,
        legend style={at={($(0,0)+(1cm,1cm)$)},legend columns=4,fill=none,draw=black,anchor=center,align=center, font=\small},
        legend to name = models]
    \addplot[color=customblue,
             mark=*,
             mark size=1] 
    coordinates{(1, 0.561) (2, 0.605) (5, 0.667) (10, 0.671) (20, 0.675) (50, 0.670) (100, 0.677)};
    \addplot[color=customyellow,
             mark=*,
             mark size=1.1] 
    coordinates{(1, 0.363) (2, 0.433) (5, 0.530) (10, 0.583) (20, 0.601) (50, 0.613) (100, 0.621)};
    \addplot[color=customred,
             mark=*,
             mark size=1.1] 
    coordinates{(1, 0.451) (2, 0.501) (5, 0.58) (10, 0.602) (20, 0.616) (50, 0.626) (100, 0.637)};
    \addlegendentry{DraftRec};    
    \addlegendentry{S-POP};
    \addlegendentry{SASRec};
    \coordinate (c2) at (rel axis cs:0,1);
    \end{groupplot}
    %------------------------------------------------------------%
    \coordinate (c3) at ($(c1)!.5!(c2)$);
    \node[above] at (c3 |- current bounding box.north)
    {\pgfplotslegendfromname{models}};
\end{tikzpicture}
\vspace*{-5mm}
\captionsetup{justification=centering}

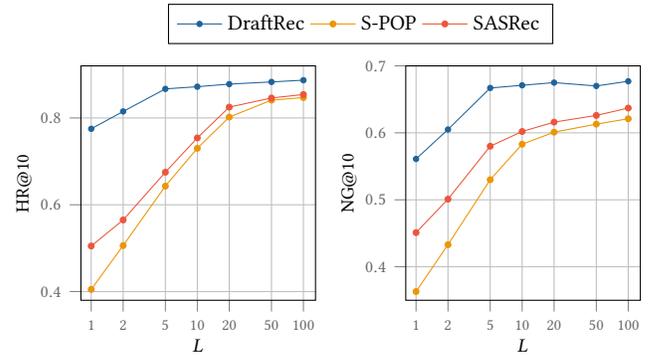
\captionof{figure}{Performance comparison of sequential recommendation models with varying history length $L$.}
\label{fig:max_seq_len_chmp_prediction}

\noindent
\\
\textbf{Champion Recommendation.} Fig. \ref{fig:max_seq_len_chmp_prediction} displays the  recommendation performances of sequential recommendation models which includes DraftRec, S-POP, and SASRec. 
We observe that player match history length $L$ is a crucial factor for the recommendation performances where both HR@10 and NG@10 increase when the player match history length $L$ increases.
However, we also discovered that both HR@10 and NG@10 converges around $L=50$ for all models. This implies that the player's champion selections are more affected by their recent matches rather than their old ones. 

%as the maximum sequence length 
%$L$ is also highly dependent
%on the average sequence length of the dataset
%with different maximum length N

\pgfplotsset{compat=1.16}
\begin{tikzpicture}
\begin{groupplot}[
      group style={group size=2 by 2, horizontal sep=1.2cm, vertical sep=1.2cm},
      width=4.7cm, height=4.7cm]
    %------------------------------------------------------------%
    %\nextgroupplot[],
    \nextgroupplot[
        every x tick label/.append style={font=\scriptsize\color{gray!80!black}},
        xmin=0.8, xmax=130,
        xtick={1,2,5,10,20,50,100},
        xticklabels={1,2,5,10,20,50,100},
        xlabel style={yshift=1mm},
        xmode=log,
        %log ticks with fixed point,
        xlabel={\footnotesize$L$},   
        every y tick label/.append style={font=\scriptsize\color{gray!80!black}},
        ymin=0.525, ymax=0.56,
        ylabel={\footnotesize ACC},
        ylabel style={yshift=-1mm},
        grid=both,
        tick align=outside,
        tick pos=left,
        legend style={at={($(0,0)+(1cm,1cm)$)},legend columns=4,fill=none,draw=black,anchor=center,align=center, font=\small},
        legend to name = models]
    \addplot[color=customblue,
             mark=*,
             mark size=1] 
    coordinates{(1, 0.538) (2, 0.542) (5, 0.545) (10, 0.549) (20, 0.552) (50, 0.554) (100, 0.553)};
    \addplot[color=customgreen,
             mark=*,
             mark size=1.1] 
    coordinates{(1, 0.529) (2, 0.531) (5, 0.538) (10, 0.540) (20, 0.541) (50, 0.5415) (100, 0.5415)};
    \addlegendentry{DraftRec};    
    \addlegendentry{OptMatch};
    %\addlegendentry{MOI-Mixer};
    \coordinate (c1) at (rel axis cs:1,1);
        \nextgroupplot[
        every x tick label/.append style={font=\scriptsize\color{gray!80!black}},
        xmin=0.8, xmax=130,
        xtick={1,2,5,10,20,50,100},
        xticklabels={1,2,5,10,20,50,100},
        xlabel style={yshift=1mm},
        xmode=log,
        %log ticks with fixed point,
        xlabel={\footnotesize$L$},   
        every y tick label/.append style={font=\scriptsize\color{gray!80!black}},
        ymin=0.479, ymax=0.501,
        ylabel={\footnotesize MAE},
        ylabel style={yshift=-1mm},
        ytick={0.480, 0.490, 0.500},
        grid=both,
        tick align=outside,
        tick pos=left,
        legend style={at={($(0,0)+(1cm,1cm)$)},legend columns=4,fill=none,draw=black,anchor=center,align=center, font=\small},
        legend to name = models]
    \addplot[color=customblue,
             mark=*,
             mark size=1] 
    coordinates{(1, 0.493) (2, 0.491) (5, 0.490) (10, 0.489) (20, 0.486) (50, 0.484) (100, 0.484)};
    \addplot[color=customgreen,
             mark=*,
             mark size=1.1] 
    coordinates{(1, 0.497) (2, 0.495) (5, 0.4945) (10, 0.494) (20, 0.493) (50, 0.494) (100, 0.494)};
    \addlegendentry{DraftRec};    
    \addlegendentry{OptMatch};
    \coordinate (c2) at (rel axis cs:0,1);
    \end{groupplot}
    %------------------------------------------------------------%
    \coordinate (c3) at ($(c1)!.5!(c2)$);
    \node[above] at (c3 |- current bounding box.north)
    {\pgfplotslegendfromname{models}};
\end{tikzpicture}
\vspace*{-5mm}
\captionsetup{justification=centering}

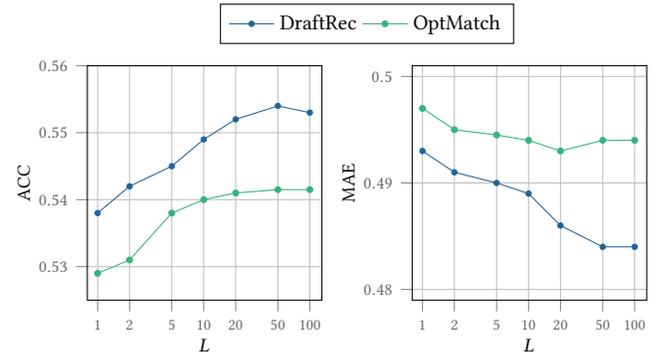
\captionof{figure}{Performance comparison of match outcome prediction models with varying history length $L$.}
\label{fig:max_seq_len_match_prediction}

\noindent
\\
\textbf{Match Outcome Prediction.} 
Fig.~\ref{fig:max_seq_len_match_prediction} shows the performance of the match outcome prediction models that utilizes player match history information which include DraftRec and OptMatch. We observed that both DraftRec and OptMatch benefits with longer player match history length $L$ and start to converge at $L=50$. When the player match history length $L$ gets longer, the models may not benefit from a longer history length since extra noise can be injected along with the extra history information.

\end{document}